\pdfoutput=1
\def\year{2018}\relax
\documentclass[letterpaper]{article} %DO NOT CHANGE THIS

\usepackage{aaai18}  %Required
\usepackage{times}  %Required
\usepackage{helvet}  %Required
\usepackage{courier}  %Required
\usepackage{url}  %Required
\usepackage{graphicx}  %Required
\usepackage{booktabs}
\usepackage[export]{adjustbox}

\usepackage{mathtools}
\usepackage{float}
\usepackage{subfig}
\usepackage{algorithmic}
\usepackage{amsmath}
\usepackage[linesnumbered,ruled]{algorithm2e}
\usepackage{multirow}
\usepackage{siunitx}

\begin{document}
% The file aaai.sty is the style file for AAAI Press 
% proceedings, working notes, and technical reports.
%
\title{Interpreting Shared Deep Learning Models via Explicable Boundary Trees}

\author{Huijun Wu$^{1,2}$, Chen Wang$^2$, Jie Yin$^2$, Kai Lu$^3$, Liming Zhu$^{1,2}$\\
$^1$The University of New South Wales, Australia\\
$^2$Data61, CSIRO\\
$^3$National University of Defense Technology, China\\
}

\maketitle

\begin{abstract}
Despite outperforming the human in many tasks, deep neural network models are also criticized for the lack of transparency and interpretability in decision making. The opaqueness results in uncertainty and low confidence when deploying such a model in model sharing scenarios, when the model is developed by a third party. For a supervised machine learning model, sharing training process including training data provides an effective way to gain trust and to better understand model predictions. However, it is not always possible to share all training data due to privacy and policy constraints. In this paper, we propose a method to disclose a small set of training data that is just sufficient for users to get the insight of a complicated model. The method constructs a boundary tree using selected training data and the tree is able to approximate the complicated model with high fidelity. We show that traversing data points in the tree gives users significantly better understanding of the model and paves the way for trustworthy model sharing. 

\end{abstract}

\section{Introduction}
Deep neural networks (DNN) have achieved great success in image classification~\cite{Krizhevsky:2017:ICD:3098997.3065386}, speech recognition~\cite{hinton2012deep} and classic games such as Go~\cite{silver2016mastering} in recent years. The success inspires the use of DNN in a rapidly increasing number of applications. Training a DNN model, however, often requires large amounts of labeled data and non-trivial efforts of tuning. Sharing a trained model is cost-effective. Machine learning models are seen offered as a service in the cloud and charged in a Pay-As-You-Go model. However, a complicated DNN model remains a black-box and the model quality is difficult to assess, particularly for a different set of data. Model accuracy and confidence values are not sufficient to reveal model behaviors on data collected from different sources. A simple example is that the confidence value of classifying a data point belonging to an unseen class by the model can be high (as shown in Fig.~\ref{fig:concept_evolution_detect} in Evaluation Section). It is challenging to make users trust a shared model.

Providing means to interpret a model is effective to improve model transparency and gain trust from model users. There has been a growing interest in exploring the interpretability for deep neural networks. Model interpretability has been tackled by different approaches in recent years. A common approach is to find examples or prototypes to guide decision making, which is along the line of the case based reasoning~\cite{aamodt1994case}. It is formulated as a set cover optimization problem to identify training data points that are close to data points within their own classes and far away from those in different classes~\cite{bien2011prototype}. The approach is complemented by~\cite{kim2016examples}, which identifies data points that do not fit a model well to further explain the model trained on data without a ``clean'' distribution.

Another common approach is to mimic a complicated model using an interpretable simple model such as decision tree~\cite{craven1996extracting,schmitz1999ann,boz2002extracting,che2016interpretable}. The limitation of this approach is that it often requires structured data. When the structure is complicated, a decision tree itself can be very complex and difficult to interpret.    
Visualization is also used to examine hidden neurons in DNNs~\cite{kabra2015understanding,yosinski2015understanding,zeiler2014visualizing}. However, in model sharing scenarios, model users may not have the access to the internal structure of a model \cite{tramer2016stealing}.

Sample perturbations~\cite{ribeiro2016should,fong2017interpretable} emerge as an important approach to understand features leading to certain predictions in image classification. These methods intend to learn an image perturbation mask that minimizes a class score. Their computational overhead is non-trivial. Moreover, they rely on the change of confidence values to infer the influence of a changing individual feature. The confidence value change can be inaccurate when there is a \emph{concept drift} in data, which can be common in model sharing scenarios. Along the line, some recent research \cite{koh2017understanding} proposes to use influence functions to find the most important training data which leads to certain predictions.  

Most of the above-mentioned approaches assume the availability of the whole training data. However, sharing the whole training data is not realistic in most of the cases due to data privacy concerns or business competition reasons. In model sharing, there is a need for a model provider to disclose a small set of training data that can help model users to gain an insight of a model to establish trust through interpretability. In this paper, we tackle this problem by providing users a small set of training data points that are enough to characterize the decision boundaries of a complicated DNN model. We employ a max-margin based approach to select the most representative training data that largely contributes to the forming of the decision boundary of the DNN model. These training data points are organized via an Explicable Boundary Tree (EB-tree) based on the distances in the DNN transformed space. The EB-tree is thus a compact structure that composes of a small set of training data but approximates the decision boundary of the DNN model with high fidelity.

Different to the example or prototype selection based approach~\cite{bien2011prototype}, EB-tree emphasizes revealing why a test data point near a boundary is classified to class A and not class B by presenting model users the traversal path of the data point among similar training data points across the DNN model decision boundary. The traversal path reveals the difference between a test data point to both sets of representative training data as well as the difference between the two sets of training data themselves, which gives a model user better interpretation of the DNN model than using examples and critics alone. Unlike perturbing based methods, EB-tree mimics the decision boundaries of the DNN classifiers by the confidence values of training points rather than perturbing the test points. This can also avoid the unexpected confidence change caused by concept drift or unseen classes~\cite{gama2014survey}. 

EB-tree is able to achieve a fidelity above 99.5\% to the DNN model with less than 0.3\% training data disclosure through a proposed~\emph{boundary stitching algorithm}. In addition, we show through a human pilot study that the traverse process of a test point in the tree clearly improves model users' understanding about how predictions are made by DNN models. We also demonstrate that in addition to giving model users insight of DNN model decision boundaries, EB-tree can be used for identifying mislabeled training data and improving the efficiency of emerging new class detection.

\section{Explicable Boundary Tree}

\subsection{Boundary Tree}
The boundary tree (forest) algorithm~\cite{mathy2015boundary} is initially proposed for fast online learning. Each node of the boundary tree represents a training point. For a query of training sample $y$, the algorithm looks up the closest node $x$ to $y$ and use the label of $x$ to predict the class of $y$. If $x$ has the same label with $y$, $y$ would be discarded. Otherwise, it is added into the tree. The process repeats for each training data point. A test data point is classified by the label of its closest node in the boundary tree. 

As each edge in the tree crosses a decision boundary, all the nodes on the boundary tree, in essence, sketches the contours of the decision boundaries. The difference between two end nodes of an edge on the boundary tree can, therefore, serve as a local explainer of the decision boundary. 

\subsection{Explicable Boundary Tree}

\begin{figure*}[h!]
\centering
\includegraphics[width=6.0in]{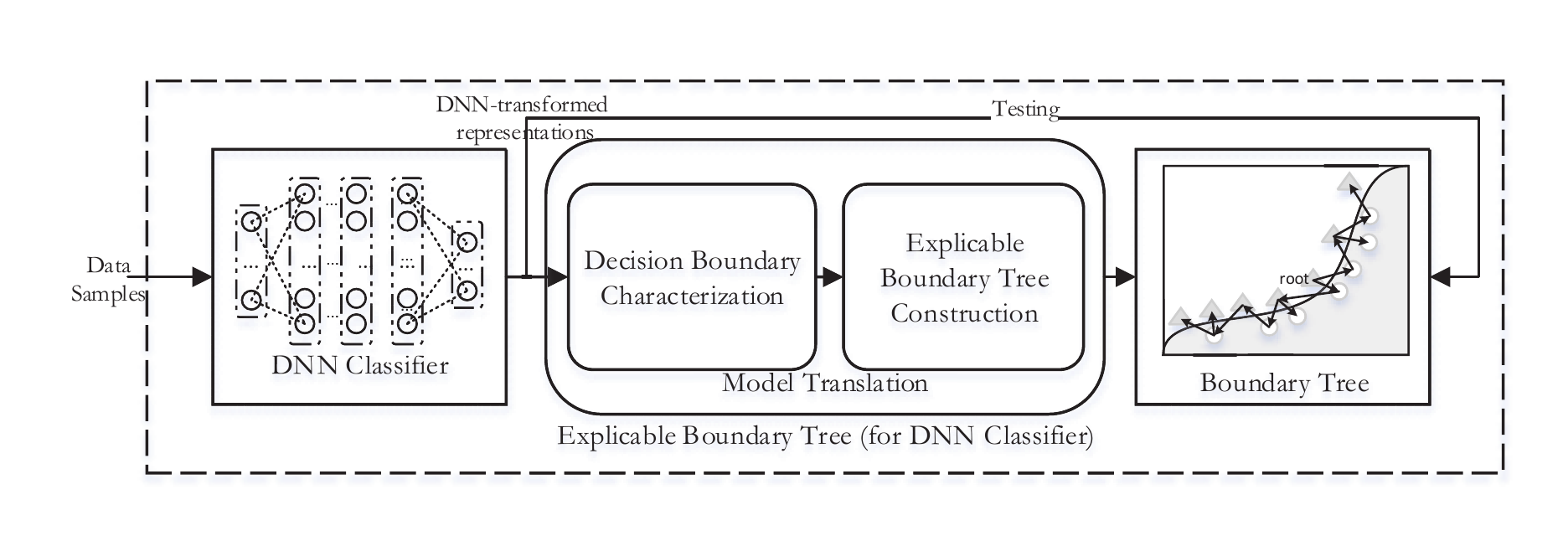}
\vspace{-10pt}
\caption{Architecture of Explicable Boundary Tree.}
\label{fig:eb_tree_arch}
\end{figure*}

Even though its edges provide certain hints about the decision boundary between two classes, a basic boundary tree has three main limitations to support interpretability of a shared DNN model: firstly, the tree has relatively low fidelity to the DNN model in decision making; secondly, the number of training data points in the tree is not optimized and there is plenty of room to reduce the number for disclosing to model users; thirdly, the training data point selection does not characterize a boundary clearly as two data points far away from the boundary may be connected by an edge. Some recent work~\cite{zoran2017learning} intends to further learning a distance metric of boundary tree edges using DNN to improve its classification accuracy and boundary representation. However, the method is not scalable due to the computing complexity. We describe EB-tree in this section to address these limitations of the basic boundary tree. Our aim is to approximate the decision boundaries of a given DNN classifier accurately with a small number of training data points. 

Figure \ref{fig:eb_tree_arch} shows the EB-tree classifier architecture. An EB-tree consists of a DNN classifier $f$ and an optimized boundary tree. The model translation module is responsible for constructing a boundary tree $T$ to mimic the decision making of the DNN classifier. Each node of $T$ is a training data point that is close to the decision boundary in the training dataset $D$. The DNN classifier transforms training data $t \in D$ to representation $f(t)$. We use the output of \emph{softmax} layer in the DNN as the transformed representation of a training data point. The distance between $f(t1)$ and $f(t2)$ is Euclidean distance.

EB-tree selects a small portion of training data points that are close to the decision boundaries to approximate a DNN model. These data points are helpful for giving model users insight about the key differences between the two classes. For a test data point, traversing the tree to reach the closest training point provides an interpretation about the decision choice of the model. 

To construct an EB-tree with good interpretability, it is important to ensure that the distance between two nodes connected by an edge is short. As an edge crosses the boundary, a short edge approximates the boundary with a narrow margin and gives a model user better visibility of the boundary. We give an ordering algorithm for training data points to achieve this boundary visibility in the following section. The algorithm also produces high accuracy and fidelity to the DNN model. 

Once constructed, the EB-tree is able to answer queries of test data points from a model user as follows: A query sample $y$ is firstly transformed by the DNN model to $f(y)$; then a traversal process locates the closest node to $f(y)$ in the tree, denoted by $x$; finally, the label of $x$ is used to predict the class of $y$. The traversal path is used as an explainer of the prediction.  

\section{Model Translation}
The model translation module in Figure~\ref{fig:eb_tree_arch} is responsible for identifying the most representative training data that characterize the decision boundaries of a DNN model and constructing an EB-tree to approximate the decision boundaries. %The goal is to provide high fidelity to the DNN model with a minimal set of training data while providing interpretability through the traversal paths in query answering.  

The basic boundary tree algorithm is not able to achieve sufficient fidelity as shown in our experiment~\ref{tab:eb_tree_dnn}, neither does it optimize the size of the tree, which may lead to unnecessary training data disclosure. We address this problem by train points sorting and boundary stitching described as below.

\subsection{Train Points Selection and Sorting}

\begin{figure}[h]
\centering
\includegraphics[width=1.8in]{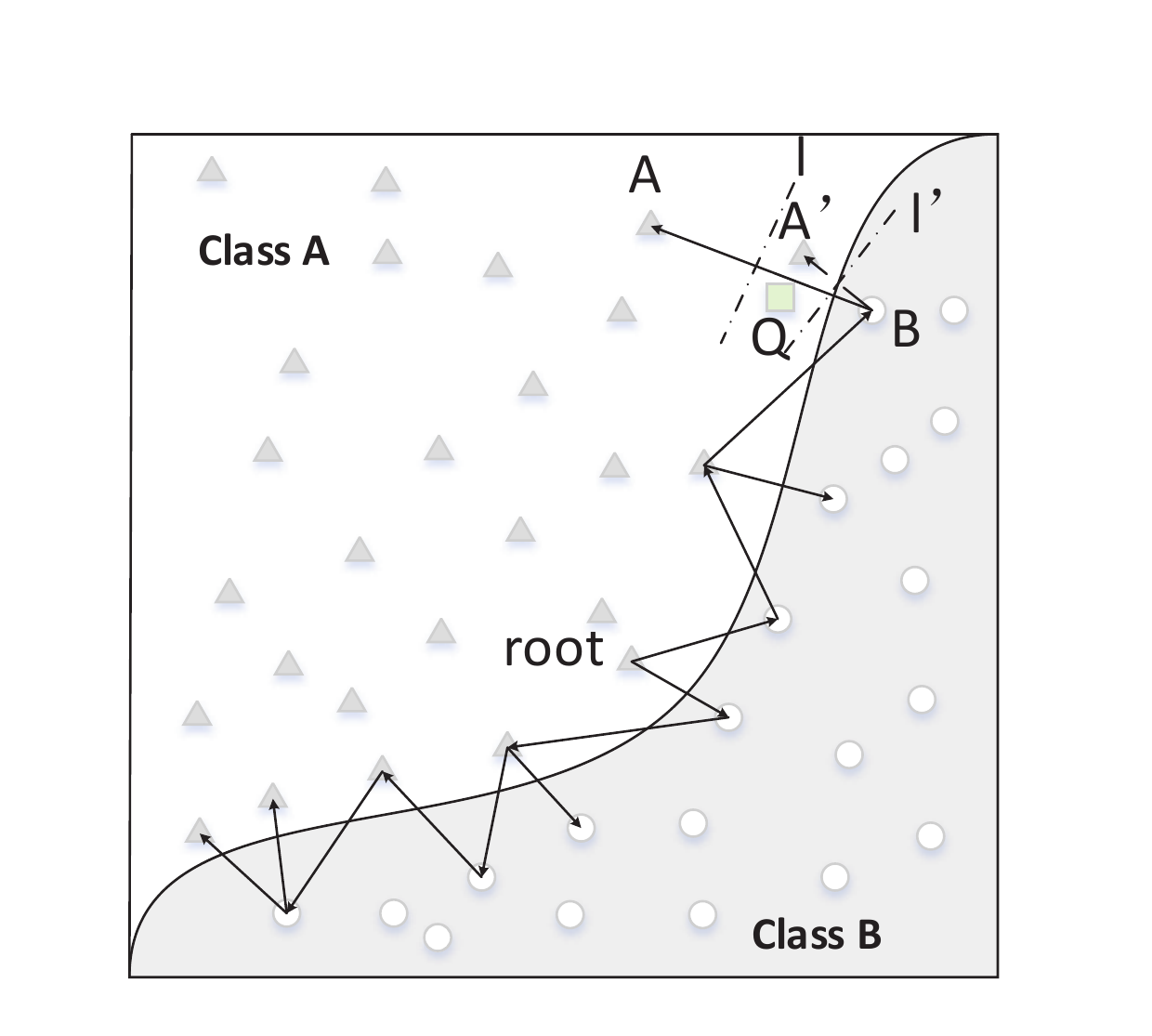}
\caption{A boundary tree for a two-class classification problem. $A$ and $A^{'}$ are two possible children for node $B$. $Q$ is a test data point.}
\label{fig:why_boundary_nodes}
\end{figure}

Figure \ref{fig:why_boundary_nodes} gives an example showing the training point selection problem. Node $B$ may add either node $A$ or $A'$ as its child. The choice leads to different classification results for query $Q$. Specifically, when $A^{'}$ is selected, query of test data point $Q$ is classified as class A because it is closer to node $A$ (on the left of dash line $l^{'}$); otherwise, it is misclassified because $Q$ is closer to node $B$ than $A$. 
 
Selecting training data pairs close to the boundary from a different side and close to each other reduces the probability a test data point is misclassified, therefore enables the constructed tree to achieve high fidelity to the DNN model. In order to do this, we need to have a metric to characterize the distance between a data point and a boundary. 

EB-tree uses a max-margin based method to measure the distance. Let $w$ denote a vector orthogonal to the decision boundary, $b$ denote a scalar ``offset'' and $\{x_{i},y_{i}\}$ denote the DNN transformed representations of train points. We consider DNN transformed representations linear separable. The decision boundary of two classes can be represented as Equation \ref{reordering_eq1}, in which $x$ is a set of training data points. 
\begin{equation} \label{reordering_eq1}
w^{T}x + b = 0
\end{equation}
The margin from the boundary to the nearest data points on each side of the boundary satisfies Equation~\ref{reordering_eq2}.
%For $C$ classes, the margin between each class $C_{k}(k = 0,1,...,C-1)$ and all the other classes is shown in Equation \ref{reordering_eq2}. 
\begin{equation} \label{reordering_eq2}
(w^{T}x_{i}+b)\cdot y_{i} \geq 1
\end{equation}

The margin between two classes is therefore:
\begin{equation} \label{reordering_eq3}
d = \frac{2}{||w||}
\end{equation}

Similar to support vector machine (SVM), we identify a boundary that maximizes the margin between two classes, which is equivalent to minimize $||w||$. 

We use the one-vs-all scheme~\cite{rifkin2004defense} to obtain a boundary for each class based on max-margin. For each training data point $(x,y)$ in the DNN transformed space, we can then compute its minimal distance to the decision boundary of its corresponding class. Training points are sorted in ascending order according to the distances to boundaries. 

The EB-tree construction process picks sorted data points to add to the tree. The order has a significant impact on both node number and interpretability of the boundary tree constructed. Consider that training points are randomly included into the tree, a training data point that is far away from a decision boundary is likely to be added to the tree first. This may result in the discard of the subsequent adjacent data points that are closer to the boundary because they share the same label. Our experiments show that random order tends to include too many nodes in the tree while achieving sub-optimal model mimicking performance. 

On the interpretability aspect, randomly ordered data points are likely to include many long edges in the boundary tree, which make the feature difference between a parent and a child node difficult to infer. In contrast, an ascending order can keep the data points near a boundary in the constructed tree, meanwhile avoiding the inclusion of long edges. It is also likely to discard data points far away from boundaries. The classification of these data points is often consistent with human intuition and including them in the tree does not contribute to boundary characterization much. This approach helps to minimize the training data disclosure to model users. 

\subsection{Boundary Stitching Algorithm}
\begin{figure}[thb]
\centering
\vspace{-20pt}
\subfloat[decision distance increasing order.]{
{\includegraphics[width=1.60in]{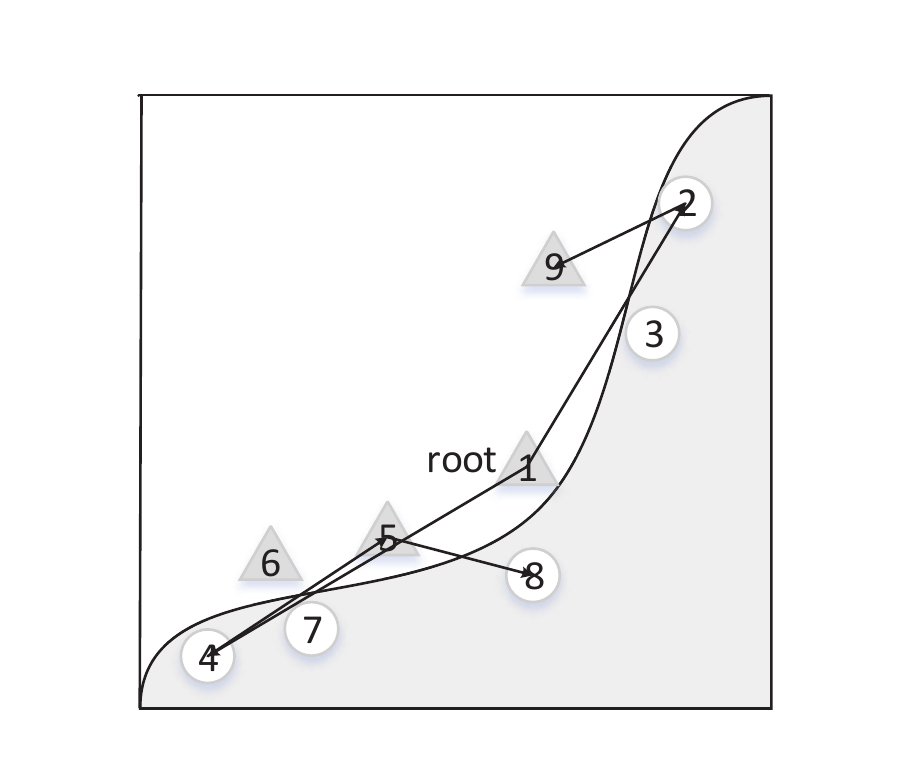}}
\label{fig:worse_bd_tree}
}
\hfill
\subfloat[boundary stitching order.]{{\includegraphics[width=1.60in]
{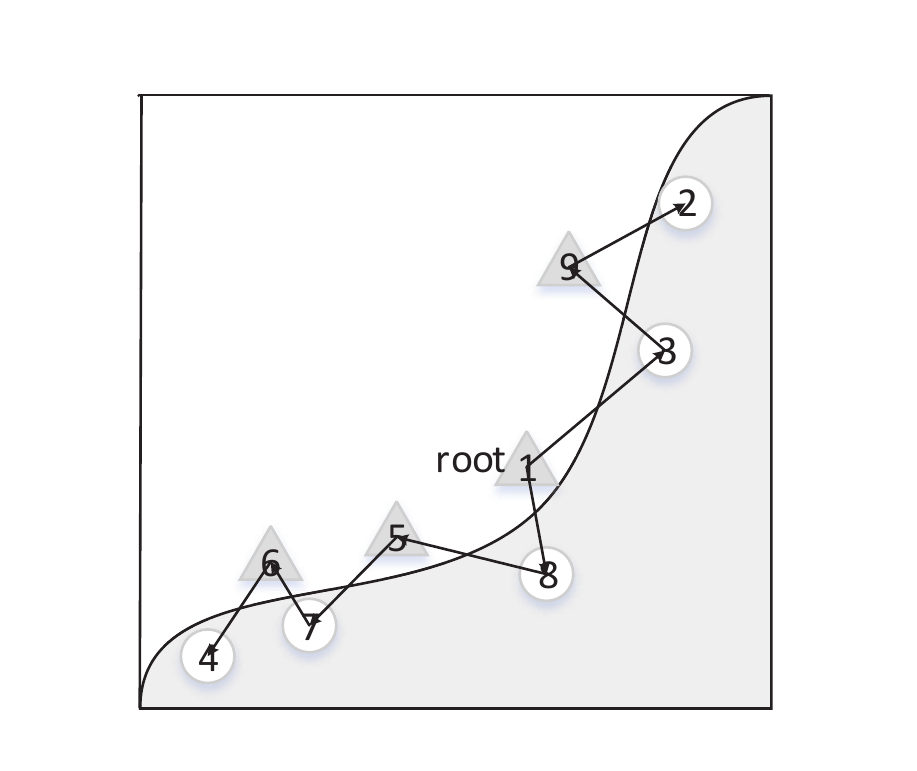}}
\label{fig:better_bd_tree}
}
\caption{The comparisons between the boundary tree simply built by decision distance increasing order and boundary stitching order. The numbers on nodes follow the increasing order of decision distance.}
\label{fig:simple_example}
\end{figure}

Simply inserting training data points according to the ascending order of their distances to boundaries is not enough to construct a boundary tree with good interpretability. It is mainly due to that two training points with similar distances to the boundary may be located at different ends of the boundary and far away from each other. The features of such two points are unlikely to share sufficient commonality for a model user to understand the decision boundary. Figure \ref{fig:worse_bd_tree} illustrates the case with an example, in which data points like node 3, 6 and 7 are not included in the tree because their corresponding closest nodes with the same label but closer to the boundary are already in the tree at the time they are being processed.  In the following, we give a boundary stitching algorithm to address this problem and the algorithm intends to construct a tree well approximating the boundary as illustrated in Figure~\ref{fig:better_bd_tree}. 

The boundary stitching algorithm shown in Algorithm \ref{alg:boundary-stitching} takes into account the distances between current node in the tree and the candidate data points to insert. It prioritizes candidate data points with the different label based on their distances to the current node in the tree. 

\begin{algorithm}[h]
\small
 \SetKwInOut{Input}{Input}
 \Input{$R$: A list of the DNN transformed representations of training points and corresponding DNN predictions.}
   \SetKwInOut{Output}{Output}
   \Output{The EB-tree $T$ for the DNN model.}
 \SetKwFunction{algo}{EB-tree Construction}\SetKwFunction{proc}{proc}
 \SetKwProg{myalg}{Procedure}{}{}
 \myalg{\algo{}}{
 //sort points according to distances to boundaries \\
 $Q$ = sortedByDistanceToBoundaries($R$)\\
 current = $Q$.removeFirst()\\
 T.insert(null, current) // insert root \\
  \While {$Q$ is not empty}{
        child = getCandidate(current,$Q$,T, k)\\
     
        // traverse T to find the closest node to child \\
        parent = findParent(T, child) \\
        \eIf{parent.label $\neq$ child.label } {
            T.insert(parent, child) \\
            current = child \\
        }

   }
}
 \caption{Boundary Stitching Algorithm}\label{alg:boundary-stitching}

  {
  \SetKwFunction{algo}{getCandidate}
  \SetKwProg{myalg}{Function}{}{}
  \myalg{\algo{$current, Q, T, k$}}{
  currentIndex = $Q$.index(current)\\
  //find $k$ nearest candidates for the head of $Q$ by LSH\\
  kNearests = LSH[currentIndex/($\frac{N}{n}$)].query(current, k)\\
  \If{kNearests is not empty}{
      expectedLabel = current.secondClosestClass()\\
      \ForEach {$n \in kNearests$}{
          \If{n.label == expectedLabel}{
            return n}
        }
    
     }
        return $Q$.removeFirst()
 }
 }
\end{algorithm}

The algorithm first computes the DNN transformed representations of training data points and their distances to the boundaries via max-margin. It then sorts these data points according to the ascending order of their distances to the boundaries. The construction of the boundary tree starts from the node with the shortest distance to a boundary. The node is inserted into the tree as the root. A search for $k$ Nearest Neighbors (kNNs) of the newly added node in the tree is then performed. If there is a data point among these neighbors falling into the second closest class of the new node, the data point is selected to insert into the tree. The insertion process traverses the selected data point to its closest node. If the data point has a different label with its closest node in the tree, it is inserted as a child of this node; otherwise, the data point is discarded as a similar node belonging to the same class has already been added to the tree. The process continues until all data points are processed. 

Finding $k$ nearest neighbors incurs high computing cost. We use \emph{locality sensitive hashing} (LSH) \cite{andoni2015optimal} to reduce the cost. The main idea of LSH is to hash the data points such that two data points close to each other have a higher probability of collision than those far away from each other. LSH can achieve a query time of $O(d \cdot n^{\rho + o(1)})$, in which $n$ is the number of data points, $d$ is the data dimension and $\rho$ controls the approximation quality. By using LSH, the complexity of the EB-tree construction process is $O(n{\log_{k} m} + n^{\rho + o(1)})$, where $m$ is the tree size. The cost is trivial compared to the cost of training a DNN model. 

\section{Evaluation}\label{sec:eval}
We evaluate the effectiveness of EB-tree on two image classification tasks. We use LeNet-5 model \cite{lecun2015lenet} for MNIST dataset and recurrent convolutional deep neural network (RCNN) \cite{liang2015recurrent} as the DNN classifier for CIFAR-10 dataset. We set $k$ in Algorithm~\ref{alg:boundary-stitching} to 32 in our experiments. We evaluate our algorithm on both model mimicking performance and the interpretability of the constructed EB-tree. The decision consistency between an EB-tree and its corresponding DNN classifier, or \emph{fidelity}, is measured by the F-measure of the EB-tree predictions against the DNN model predictions. We measure the interpretability of EB-trees by conducting a human pilot study. 

\subsection{Model Accuracy and Fidelity}

\begin{table}[]
\label{tab:eb_tree_dnn}
\centering
\scriptsize
\caption{Comparison of boundary tree, EB-tree and the original DNN classifiers. $SD$ is standard deviation.}
\begin{tabular}{@{}lccc@{}}
\toprule
    Model & Error Rate & Fidelity & \# of Nodes \\ \midrule
    MNIST-LeNet-5 & 0.56\%  &   100\%          &        N/A     \\
    MNIST-Boundary Tree & 1.02\%(SD=0.14\%) & 99.39(SD=0.11\%) & 263(SD=39)  \\
    MNIST-EB-tree     &   \textbf{0.59\%}  & \textbf{99.72\%}  & \textbf{102} \\

\midrule
   CIFAR-RCNN &    7.10\%      &    100\%         &      N/A       \\
   CIFAR-Boundary Tree &  8.54\%(SD=0.31\%)  & 98.89\%(SD=0.19\%)  & 392(SD=42)  \\
   CIFAR-EB-tree &  \textbf{7.23\%}   & \textbf{99.56\%}  &  \textbf{119}   \\
\bottomrule
\end{tabular}
\end{table}

As shown by the error rate in Table \ref{tab:eb_tree_dnn}, EB-tree can achieve an accuracy highly similar to that of the DNN classifiers. EB-tree also approximates the DNN classifiers with high fidelity (99.72\% with MNIST-LeNet-5 and 99.56\% with CIFAR-RCNN). EB-tree significantly outperforms the original boundary tree algorithm in accuracy, fidelity and node number in the tree.  Note, for the MNIST dataset, the resulting EB-tree only needs to disclose 102 out of the 60,000 training data points (0.17\%) to model users for them to understand the model. For CIFAR-10 dataset, the constructed EB-tree only needs to disclose 119 out of the 50,000 training data points (0.24\%) to model users. Further tradeoffs can be made in the data point selection process if each data point is associated with different privacy information. 

\begin{figure}[thb]
\centering
\subfloat[boundary tree.]{{\includegraphics[width=1.5in]{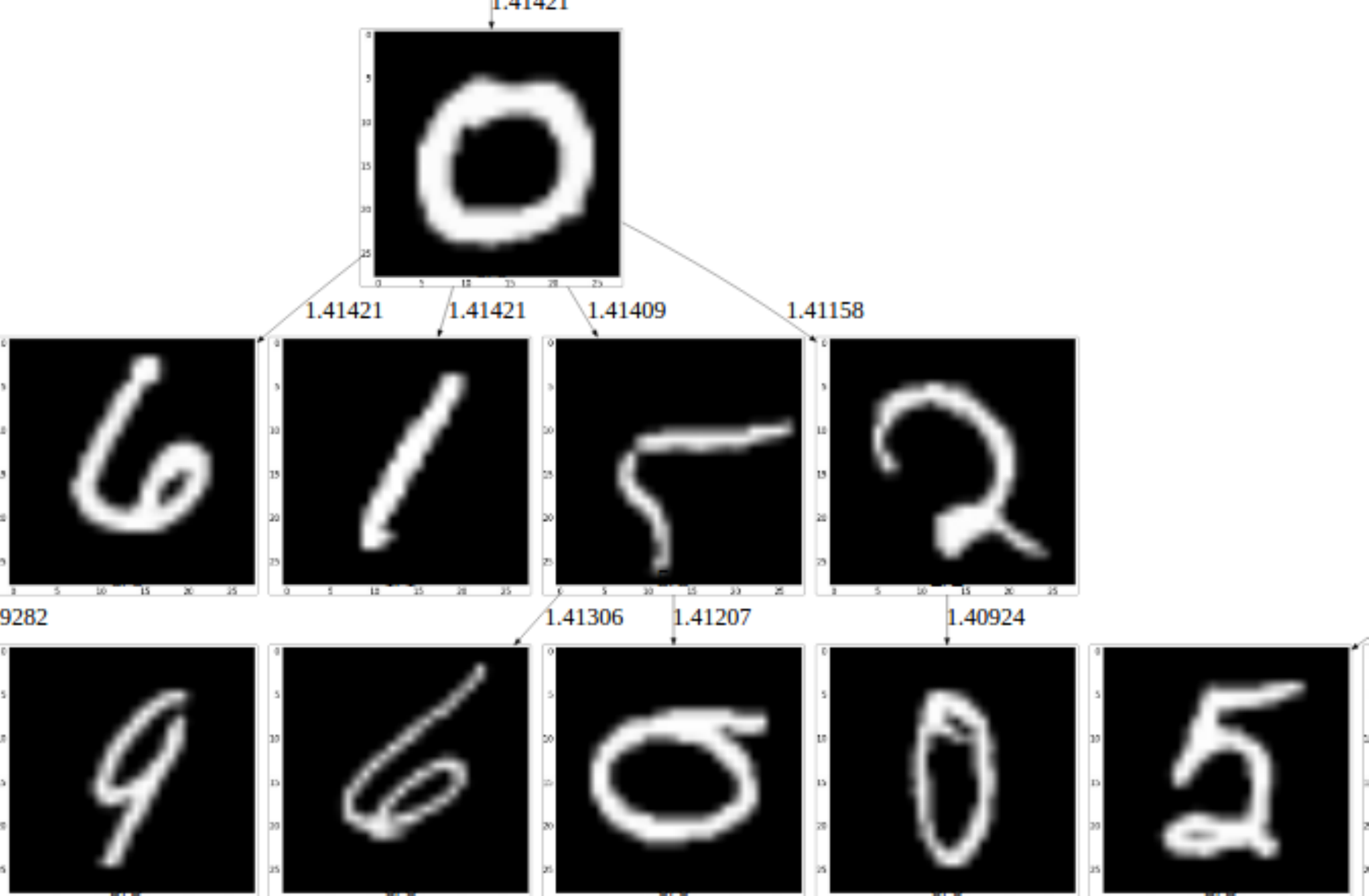}}
\label{fig:random_boundary_tree}}
\hfill
\subfloat[EB-tree.]{{\includegraphics[width=1.5in]{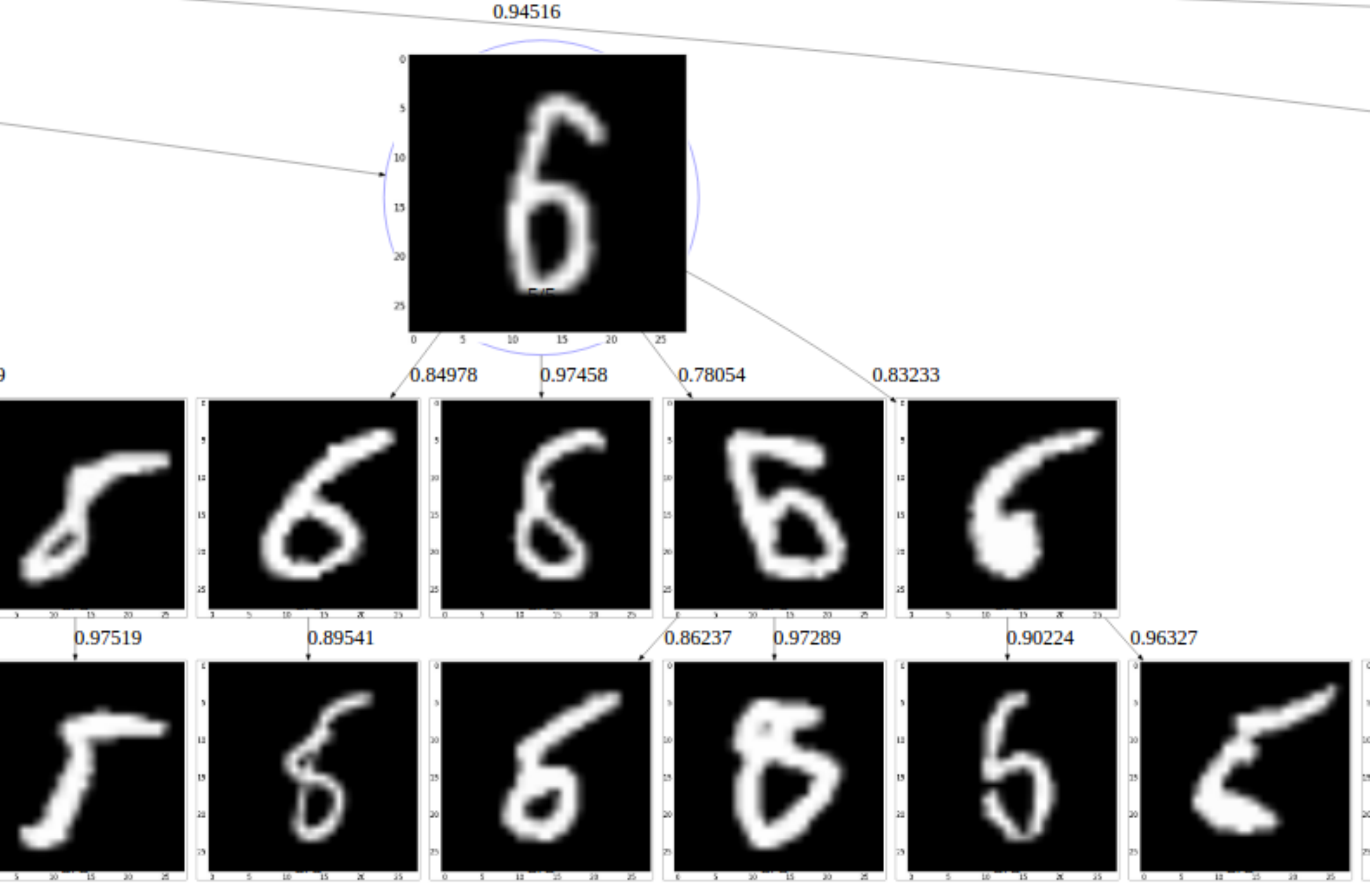}}
\label{eb_tree}}
\caption{Samples of the nodes on boundary tree and EB-tree.}
\label{fig:tree_samples}
\end{figure}

EB-tree optimizes its edges to align better with the boundary than the basic boundary tree. The effect is shown in Figure \ref{fig:tree_samples}. The nodes connected by the same edge in a basic boundary tree are visually more different than those in an EB-tree. The node selection process in EB-tree is able to identify training points with subtle difference to characterize the decision boundaries. When a test data point traverses the tree, these subtle differences give a model user insight about why the test point is classified to a specific class. 
As an example, Figure \ref{eb_tree} shows the decision boundary between some 6/8 and 6/5 pairs which give a user hints about subtle features that differentiate the classes. 

\subsection{Use Cases of EB-tree}
\subsubsection{Case 1: Explaining Predictions of Individual Data Points. }
Unlike some existing work which tries to give the closest prototype as an explanation for a given test sample, EB-tree gives the traversal path on the tree as an explanation which assists the model users to understand the reasons behind predictions.

To evaluate how effective an EB-tree is on helping a user understanding the decision making of a model, we conduct a human pilot study. The study involves 10 users without machine learning research and development experience. We design tasks to measure whether a user gains insight of a model with the help of an EB-tree.

\begin{table}[]
\scriptsize
\centering
\caption{Average traversal paths agreement ratio with human subjects.}
\label{tab:human_evaluation}
\begin{tabular}{@{}lll@{}}
\toprule
DNN Model              & Agreement ratio  \\ \midrule   
MNIST (LeNet-5)  &   97\% (SD = 2\%)     \\
CIFAR-10 (RCNN) &  96\% (SD = 3\%)   \\ \bottomrule
\end{tabular}
\end{table}

We present MNIST EB-tree and CIFAR-10 EB-tree to 10 users. We then randomly choose 100 images with 10 from each class from the test dataset of each model to show users. For each image, we provide 10 traversal paths and ask the participants to choose the one that gives them the most insight to understand the model decisions. The 10 traversal paths consist the following three categories:
\begin{enumerate}
\item The path produced by the EB-tree algorithm;
\item Two paths modified slightly from the EB-tree path;
\item Seven paths generated by random walking on the EB-tree starting from the root. 
\end{enumerate}
The result is shown in Table  \ref{tab:human_evaluation}. The agreement ratio between each user's choices and EB-tree paths is high, which confirms that EB-tree paths give users better understanding of both MNIST and CIFAR-10 DNN model. 

\begin{figure}[thb]
\centering
\subfloat[]{{\includegraphics[width=0.50in]{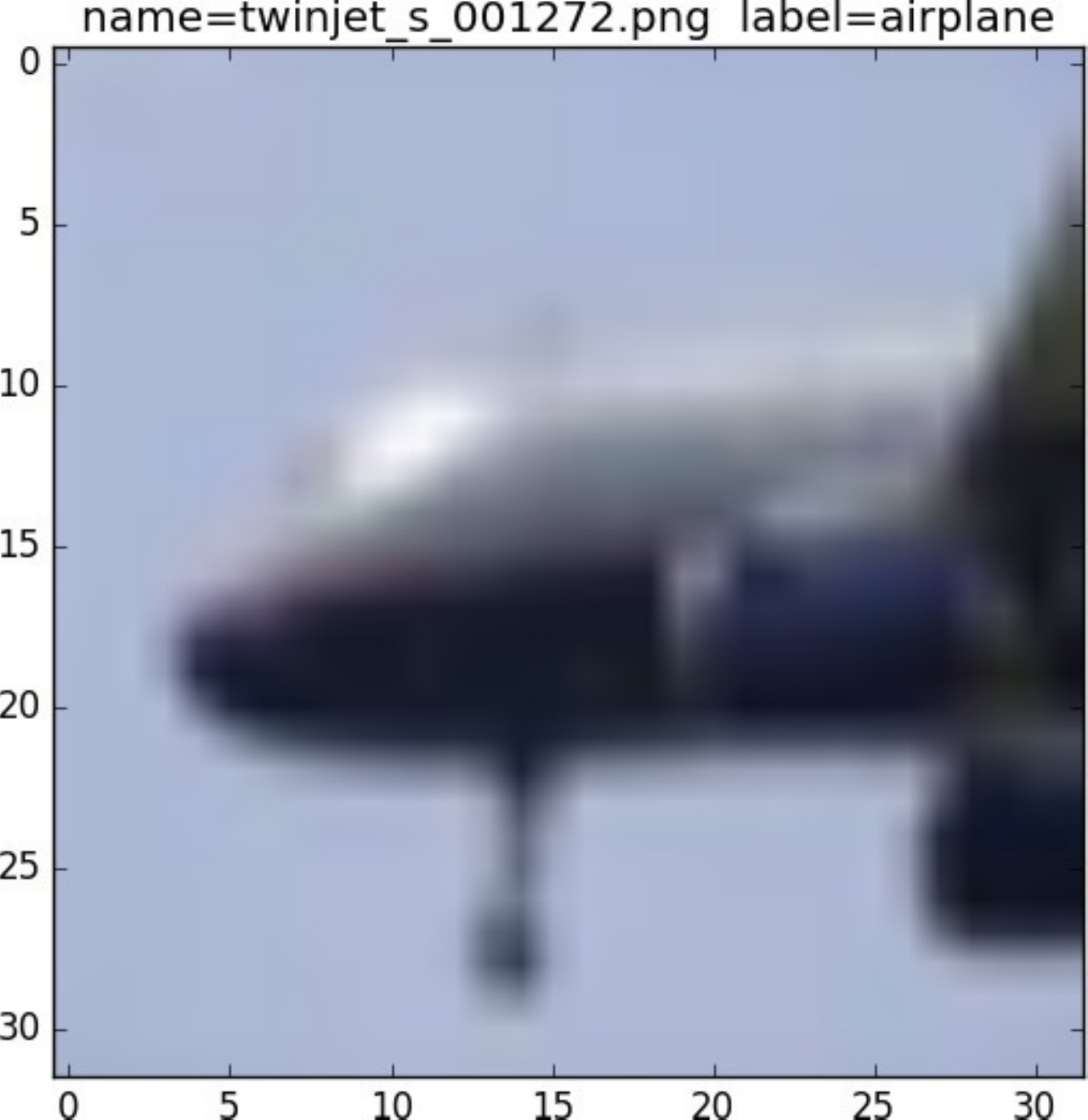}}
\label{fig:cifar_602}}
\hfill
\subfloat[]{{\includegraphics[width=0.50in]{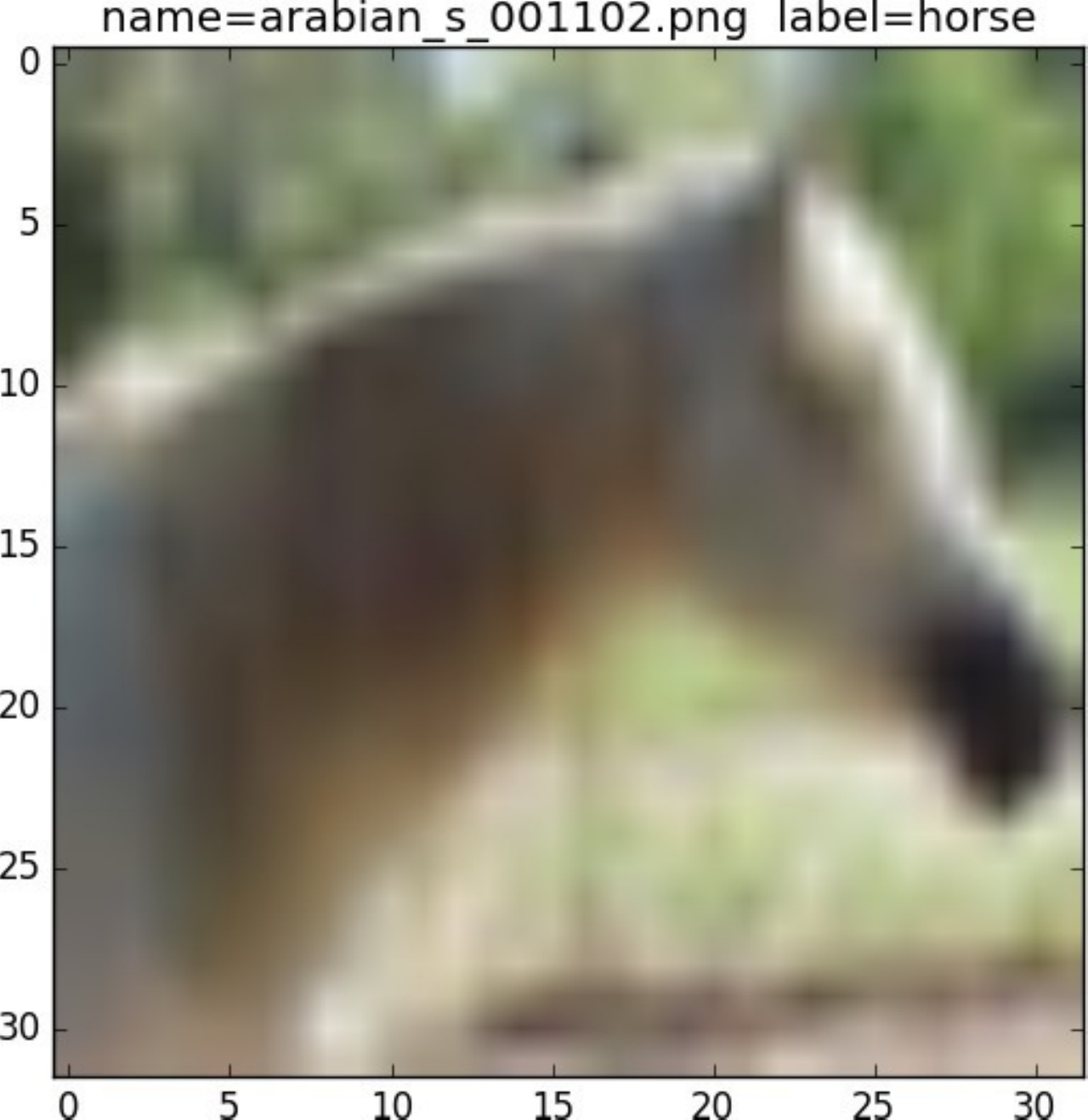}}
\label{fig:cifar_798}}
\hfill
\subfloat[]{{\includegraphics[width=0.50in]{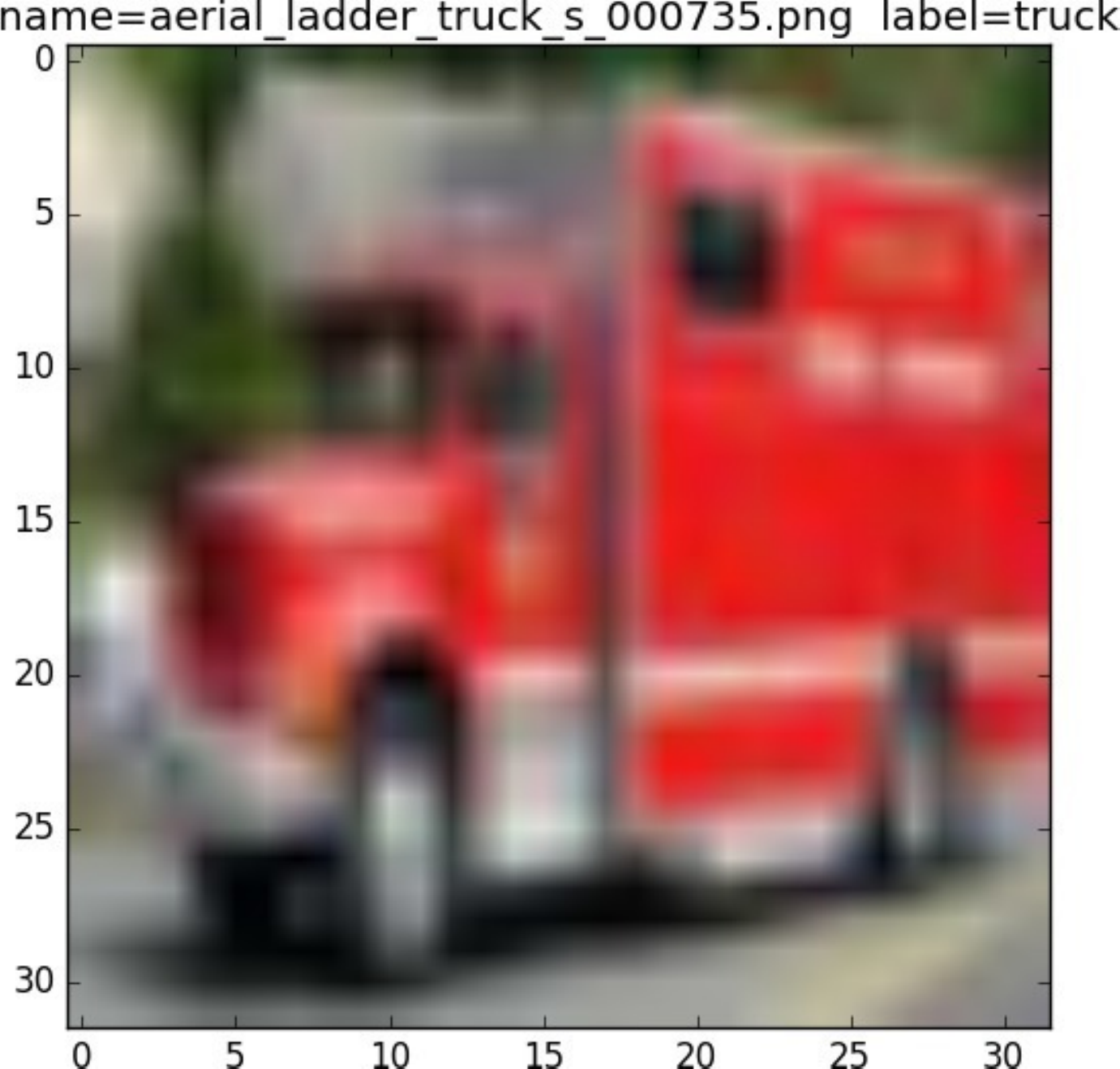}}
\label{fig:cifar_940}}
\hfill
\subfloat[]{{\includegraphics[width=0.50in]{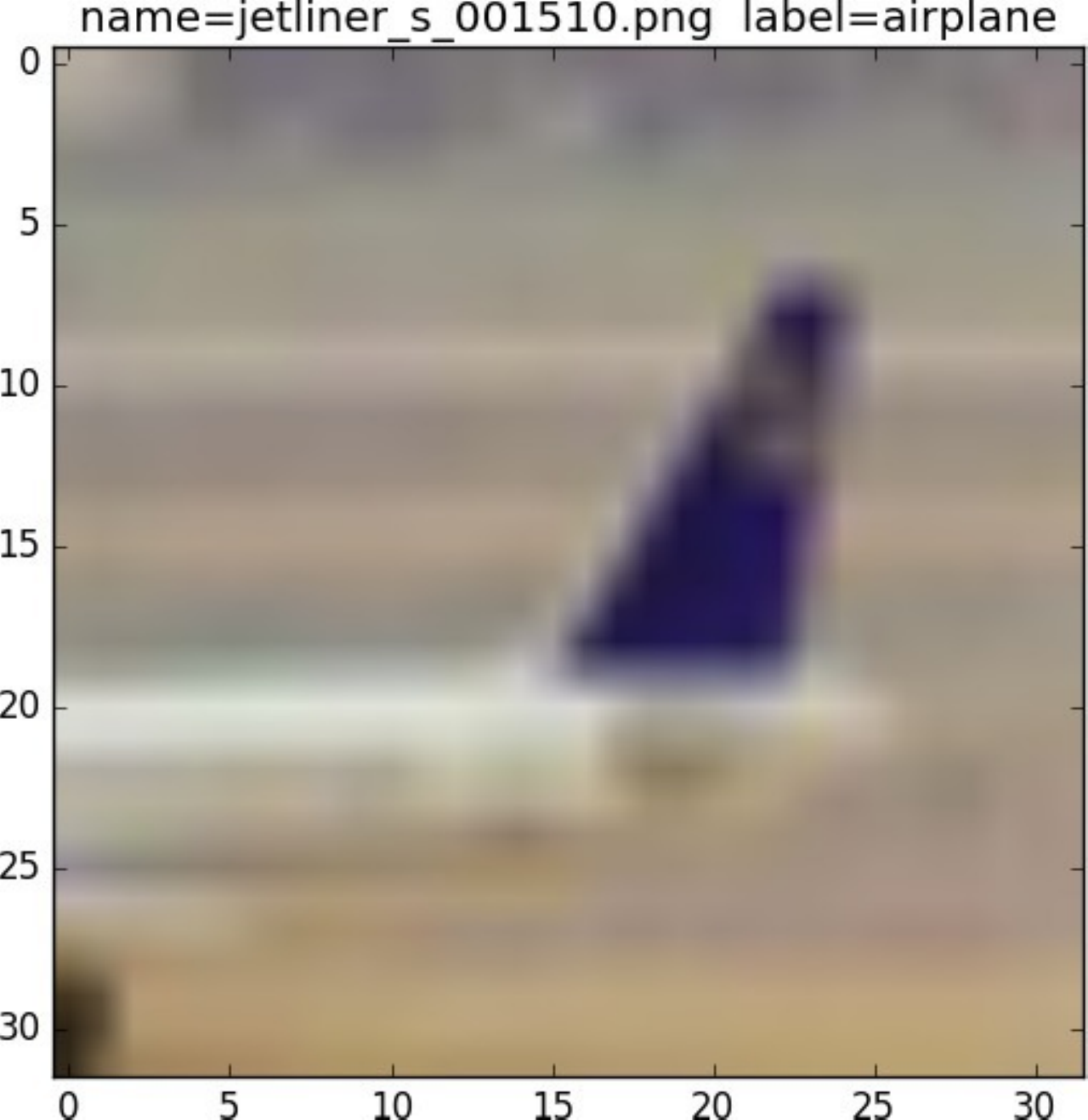}}
\label{fig:cifar_6653}}
\hfill
\subfloat[]{{\includegraphics[width=0.50in]{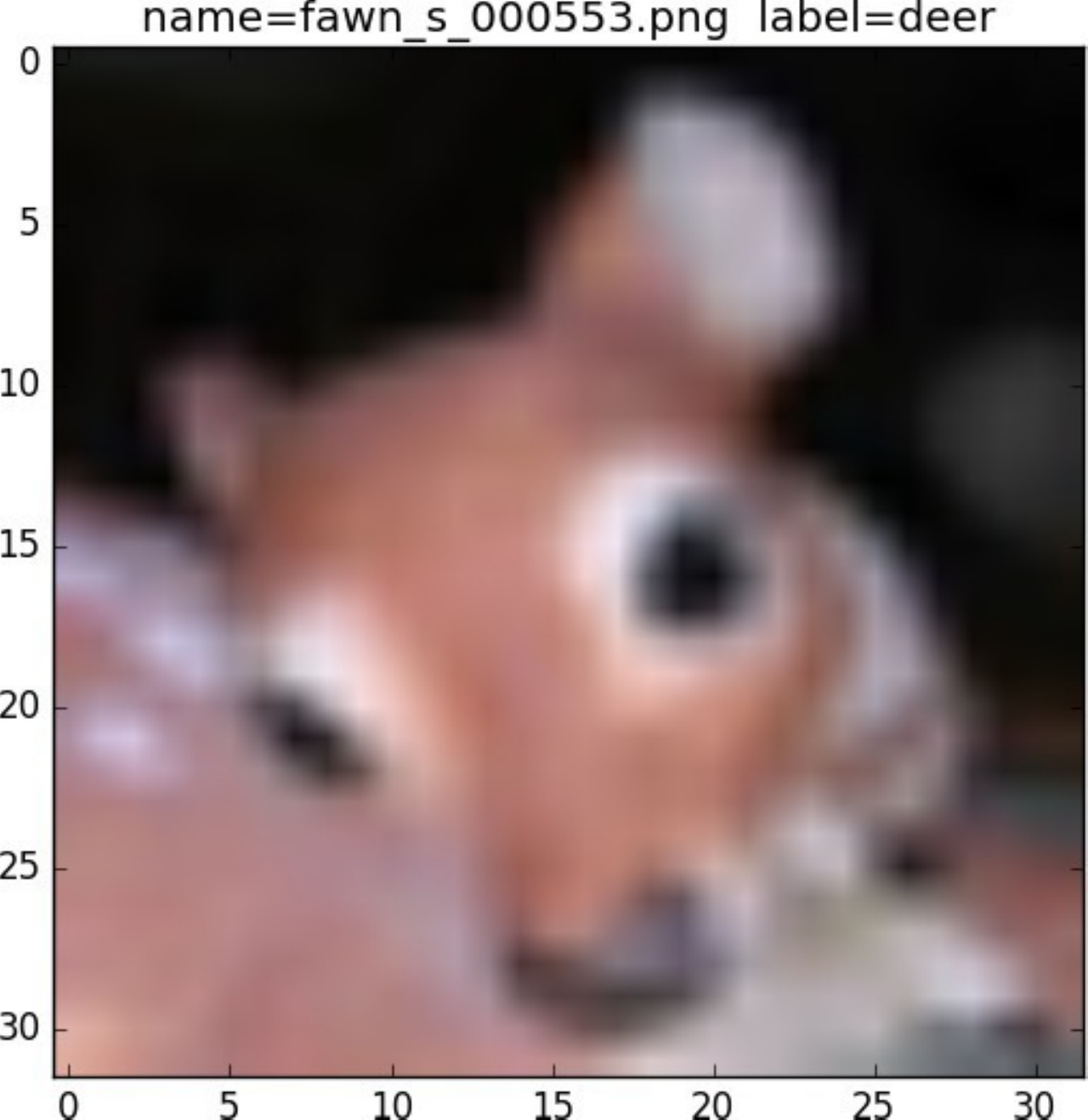}}
\label{fig:cifar_4112}}
\hfill
\subfloat[]{{\includegraphics[width=0.50in]{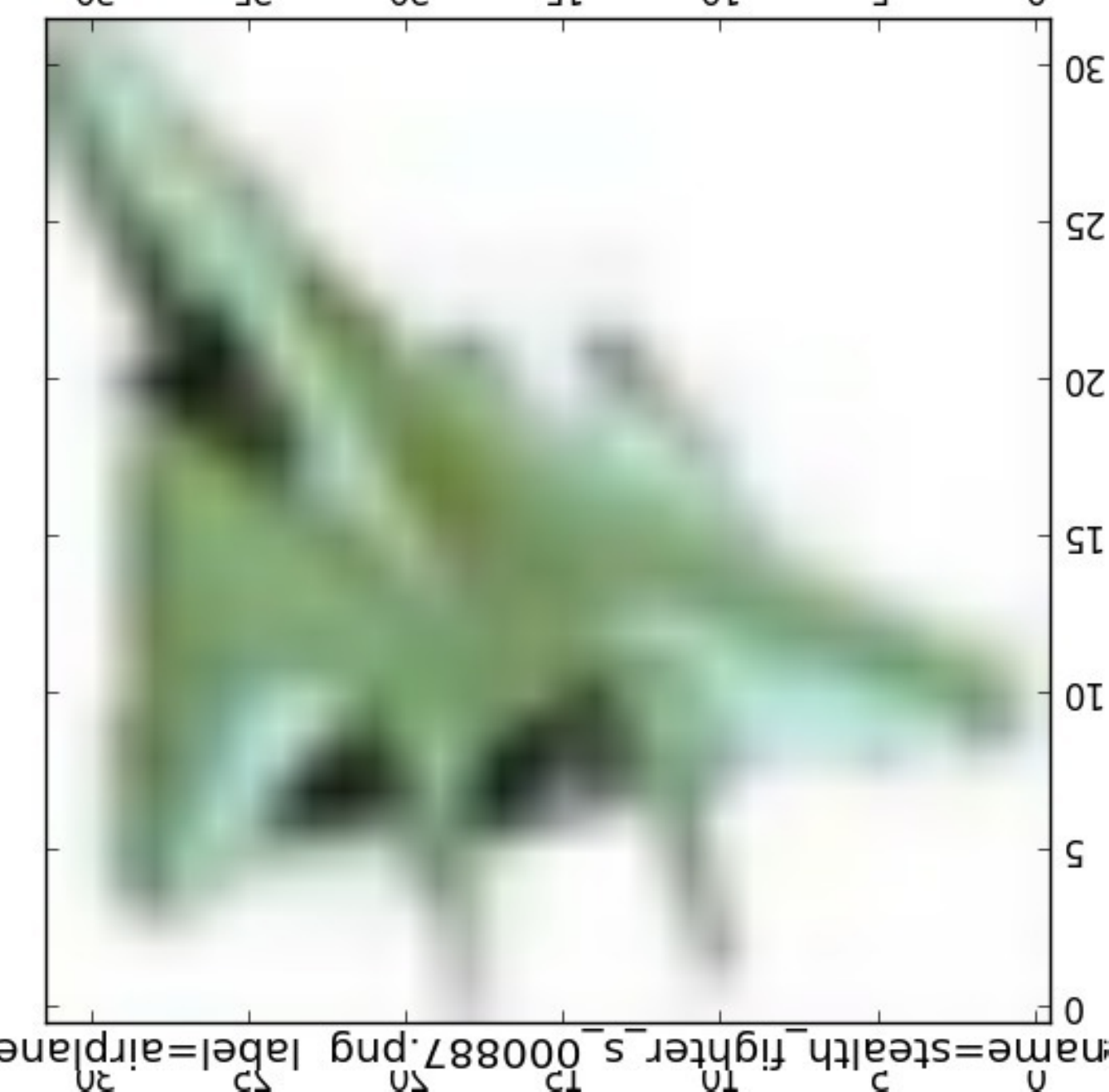}}
\label{fig:cifar_9021}}
\caption{Example test samples in MNIST and CIFAR-10 dataset.}
\label{fig:selected_test_samples}
\end{figure}

\begin{figure}[thb]
\centering
\subfloat[]{{\includegraphics[width=0.3\linewidth]{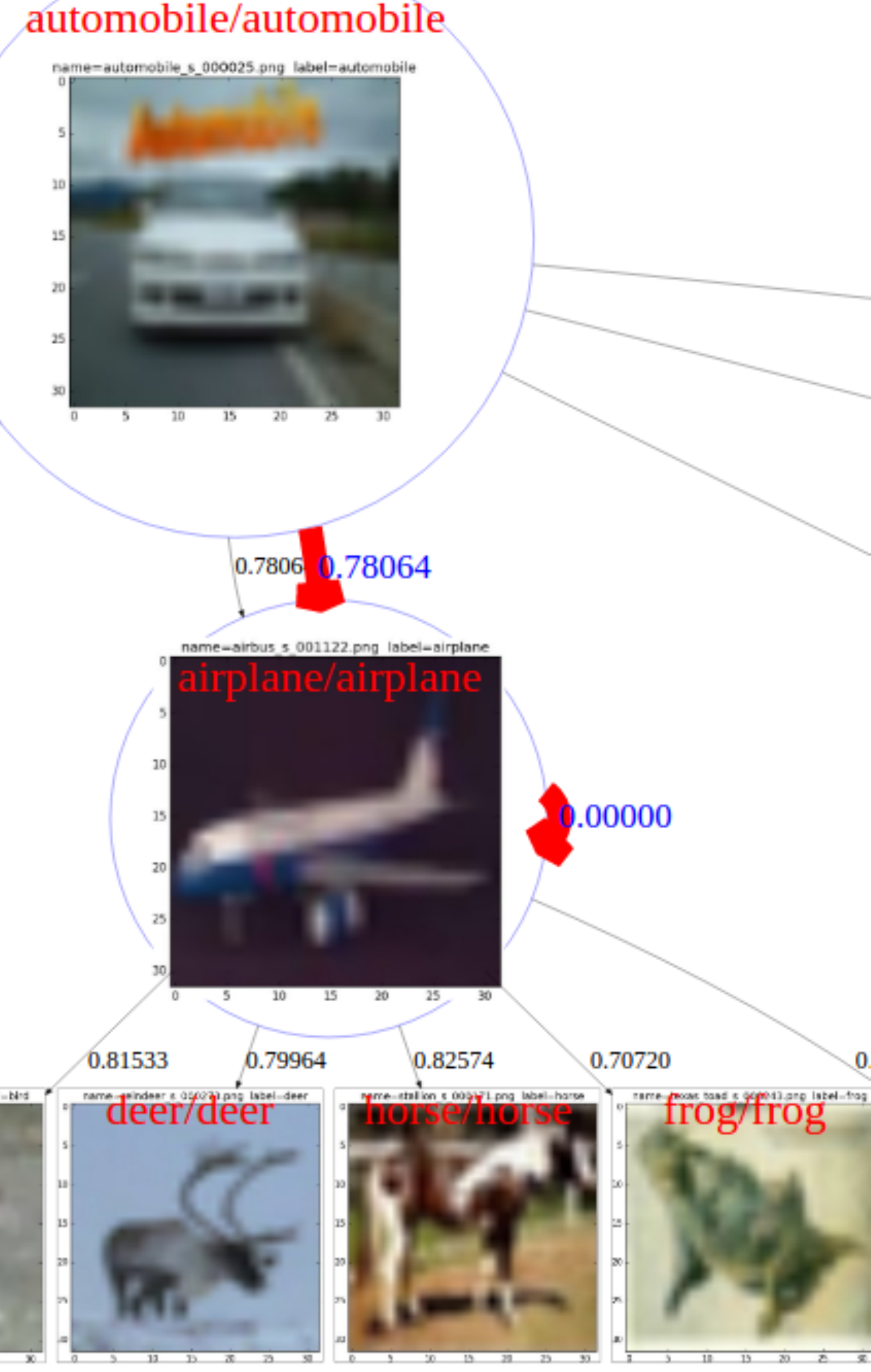}}
\label{fig:602_path}}
\hfill
\subfloat[]{{\includegraphics[width=0.3\linewidth]{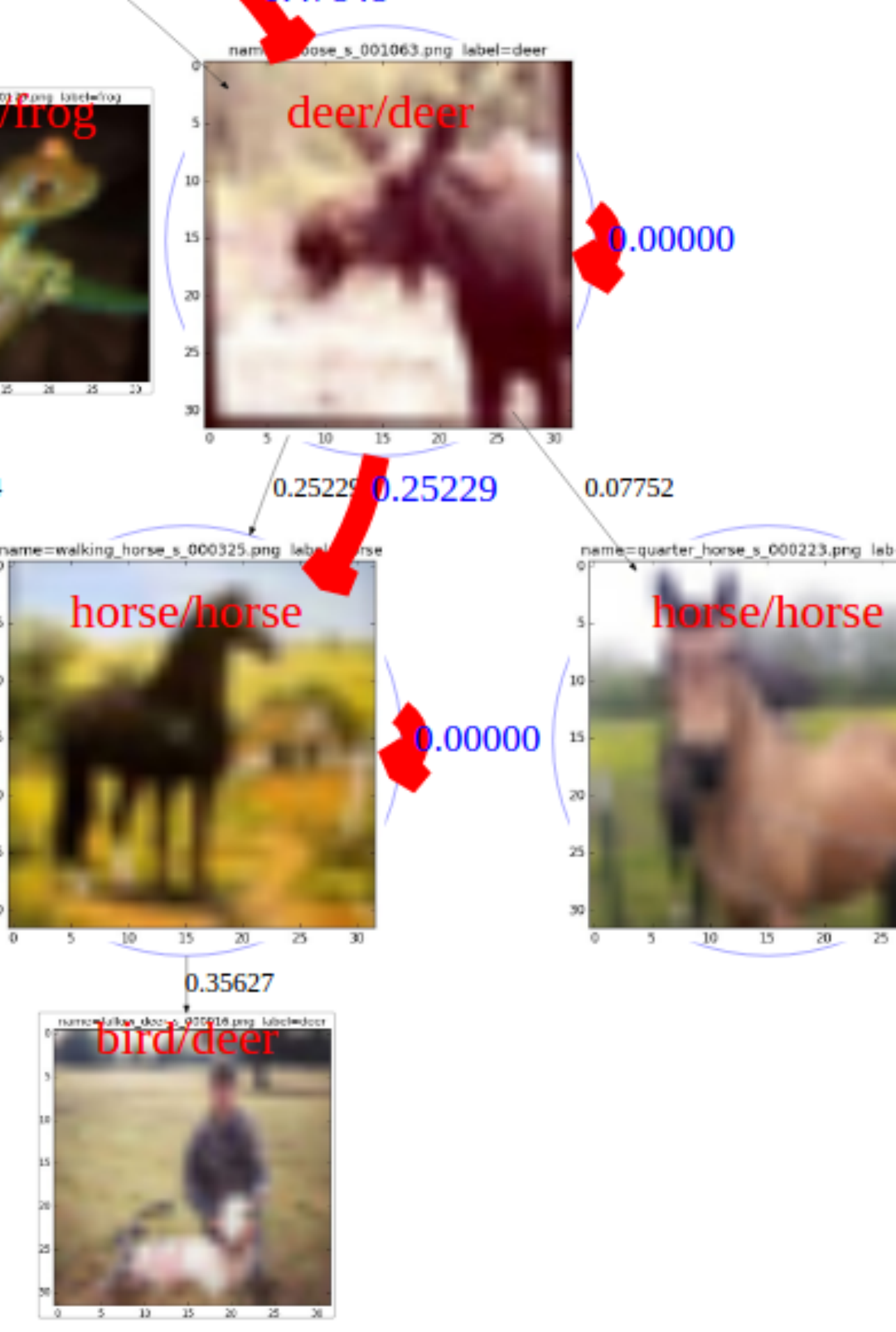}}
\label{fig:798_path}}
\hfill
\subfloat[]{{\includegraphics[width=0.28\linewidth]{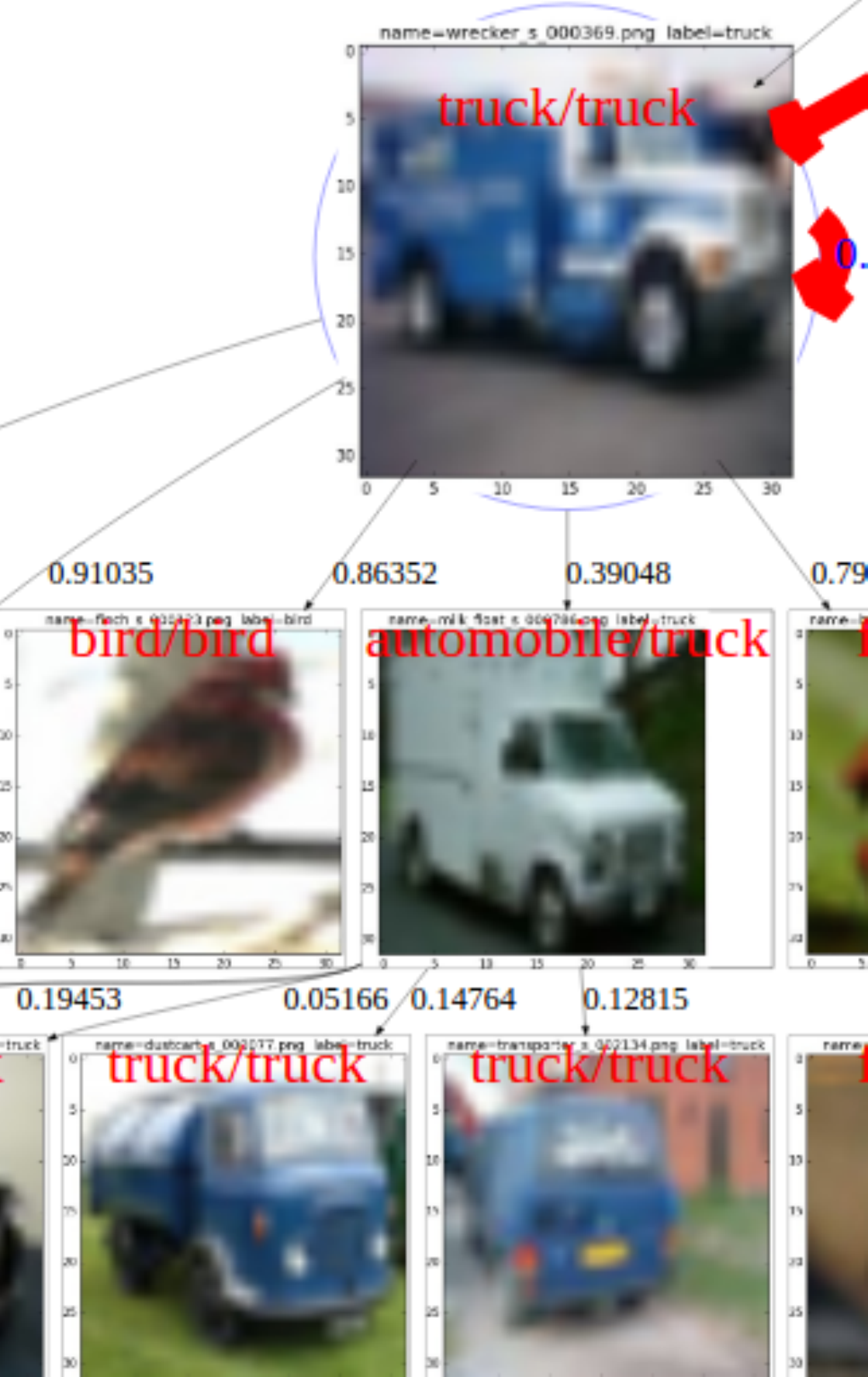}}
\label{fig:940_path}}
\vfill
\subfloat[]{{\includegraphics[width=0.3\linewidth]{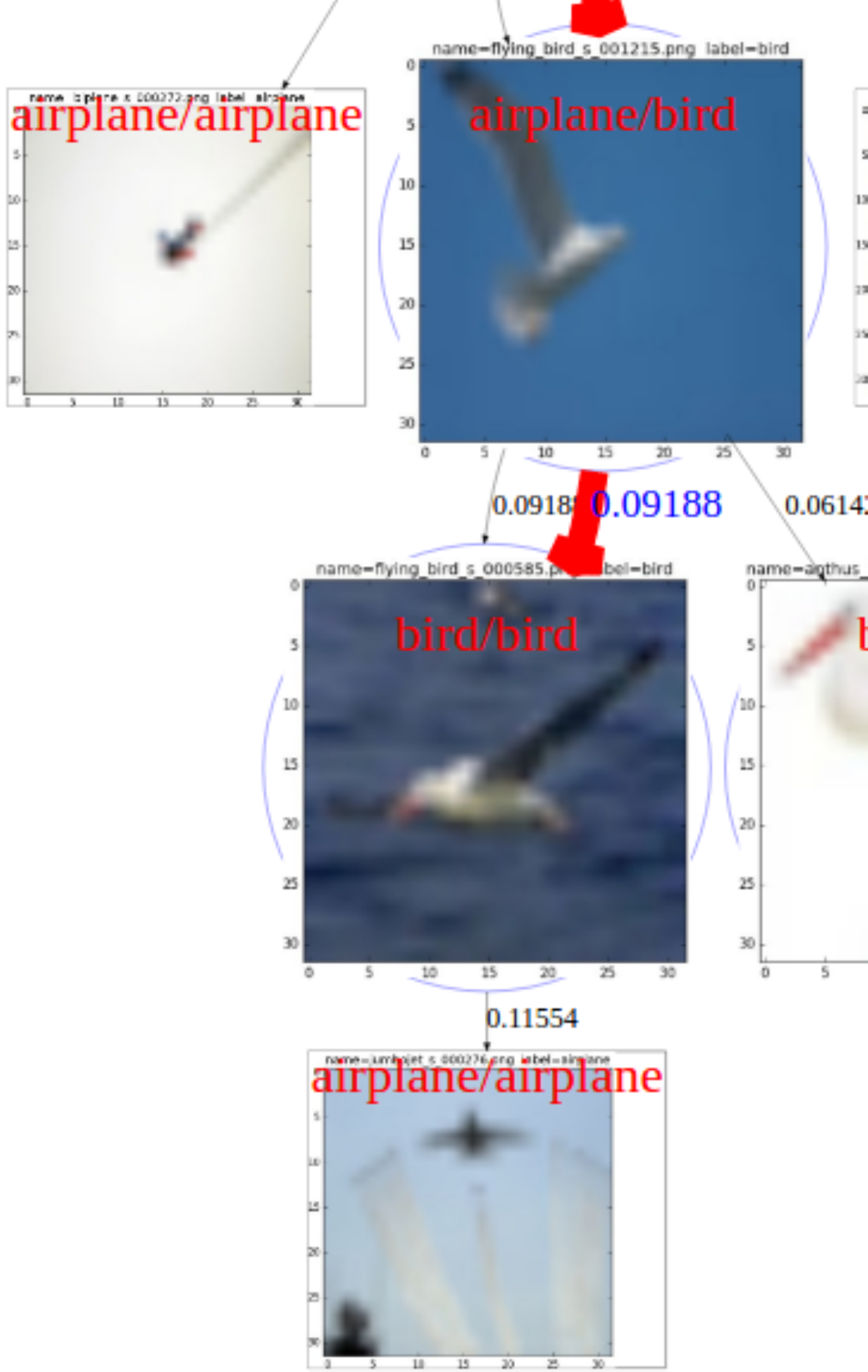}}
\label{fig:6653_path}}
\hfill
\subfloat[]{{\includegraphics[width=0.25\linewidth]{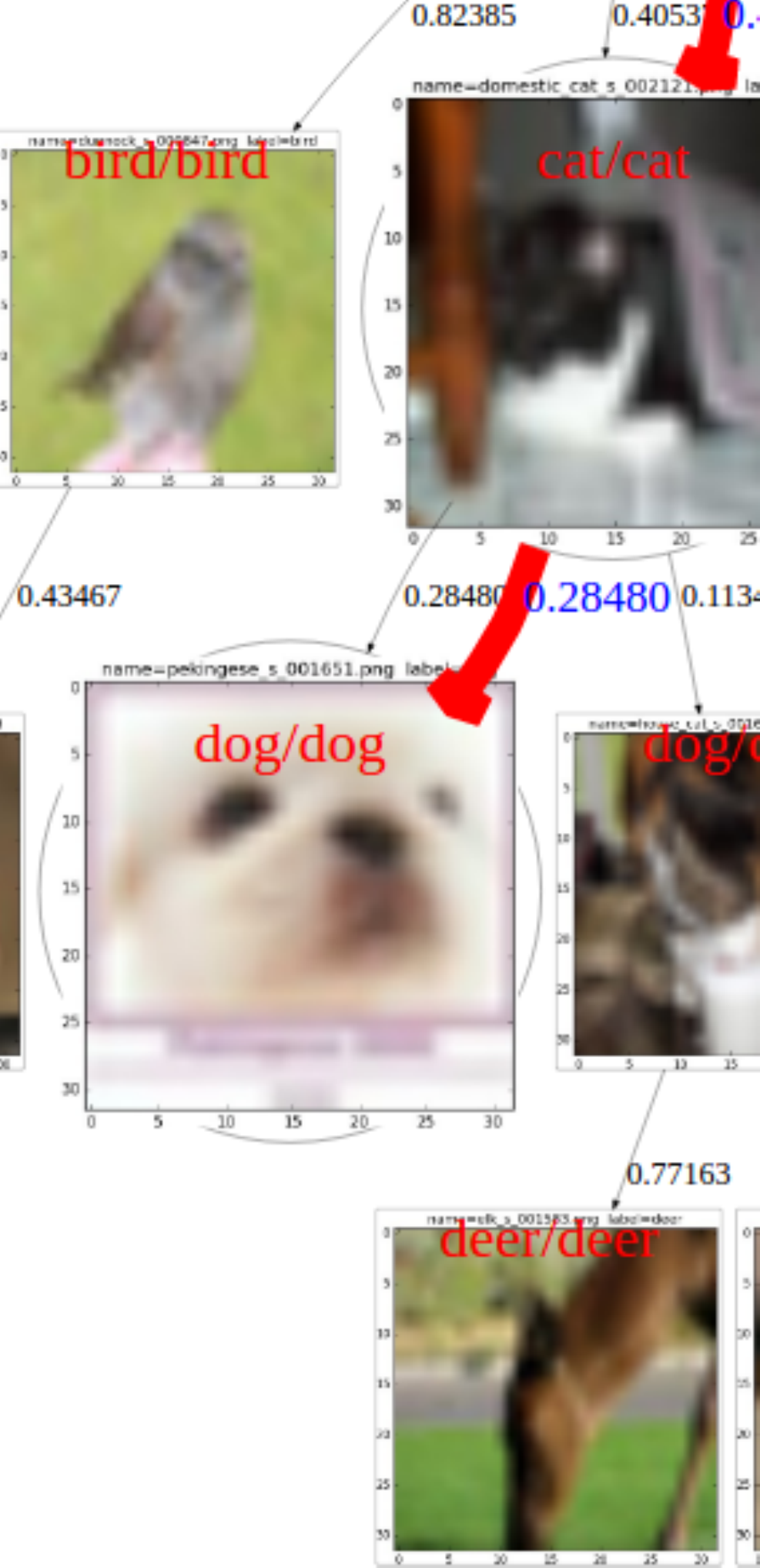}}
\label{fig:4112_path}}
\hfill
\subfloat[]{{\includegraphics[width=0.30\linewidth]{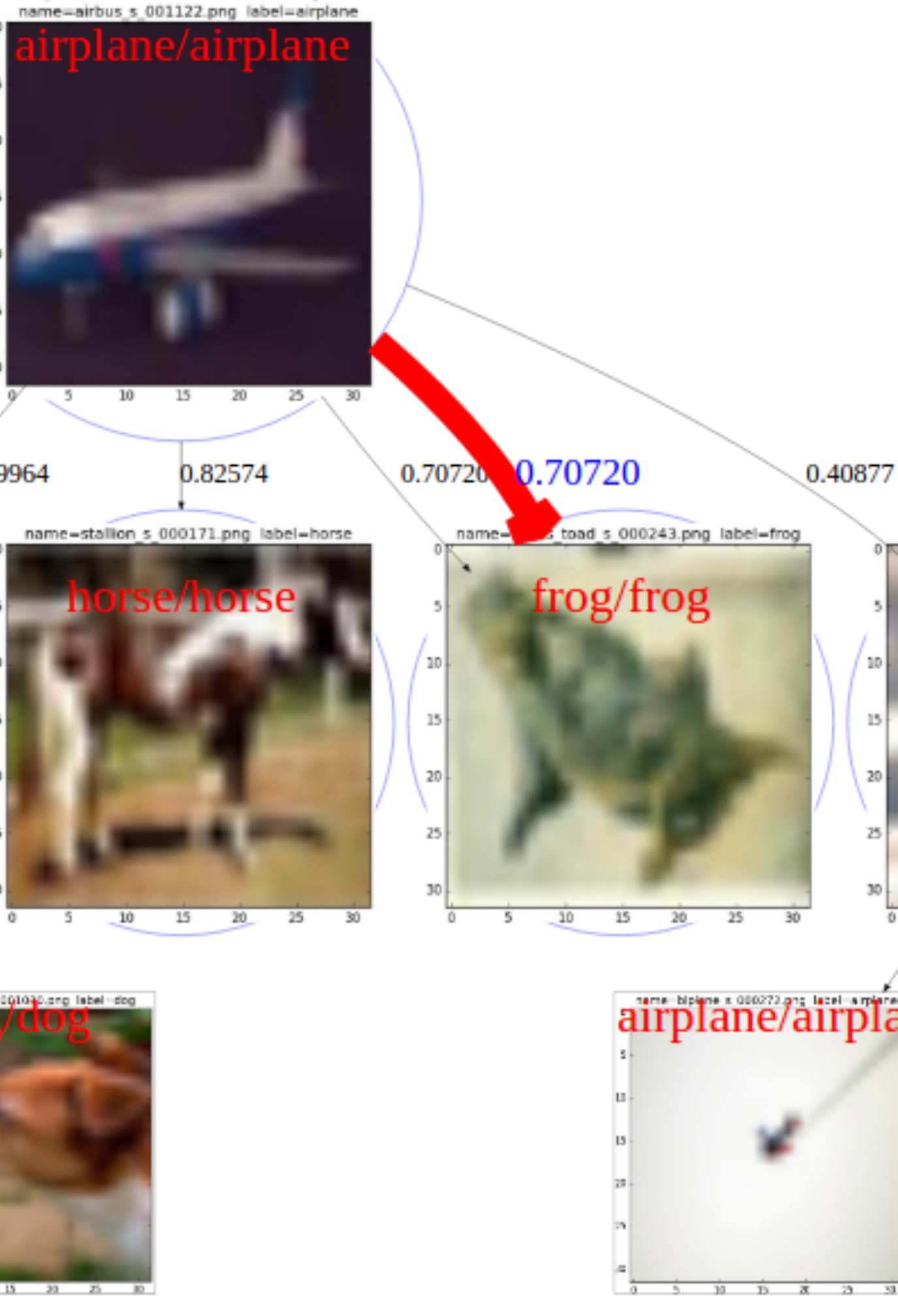}}
\label{fig:9021_path}}
\caption{Traversal paths (represented by red lines) on the EB-tree for example test images. }
\label{fig:selected_test_samples_traverse}
\end{figure}

The traversal paths and closest nodes given by the EB-tree help the model users understand some predictions in depth. Particularly they help model users notice certain features that contribute to the classification of a test data point to a specific label. For example, Figure \ref{fig:selected_test_samples} shows six test images in CIFAR-10 dataset. Among the images, the first three are correctly classified while the last three are incorrectly classified by the DNN classifier. Figure \ref{fig:selected_test_samples_traverse} shows the traversal paths of these images on the EB-tree to interpret the predictions. For the correct predictions, the label of the final node on the traversal path is used to make the prediction. In other words, the final node is considered to be more similar to the test data compared to the node's parent and children. If this 'similar' relationship makes sense to human, the model decision is consistent with human intuition; otherwise, the model decision needs a careful check. We find the traversal paths of incorrect predictions unusual. The airplane in Figure~\ref{fig:cifar_6653} is classified as 'bird', the traversal path explains this by the final node in Figure \ref{fig:6653_path}. In fact, the tail of the airplane makes it similar to a bird with the long tail in the training data. We identify that there lacks similar airplane images in the training data. The lack of similar training data contributes to the misclassification of the airplane to a bird. Similarly, Figure~\ref{fig:cifar_4112} is a deer while being classified as a 'dog' by the model. Our experiments on other test images also show a partial view of an object often leads to incorrect predictions on the CIFAR dataset. The green airplane in Figure~\ref{fig:cifar_9021} is classified as a 'frog' due to its visually similar color and shape with the frog, as shown in Figure~\ref{fig:9021_path}. Apparently, the lack of similar training airplane examples contributes to this misclassification.  

The results indicate that EB-tree is able to give model users clear hints about why a data point is classified into a particular class. A model user can traverse their own data on the tree to gain insight about the model. For model providers, EB-tree is also helpful for debugging their models. 

\subsubsection{Case 2: Visualizing Decision Boundaries.}
The decision boundaries learned by complex models like DNN classifiers are difficult to ``see'' by a human without examples. EB-tree provides a means for model users to see a small number of training samples that characterize the boundaries. It is also helpful for improving the model training. 

We implement an operation called \textit{boundary projection} on the EB-tree to visualize boundaries. This operation traverses the EB-tree and finds all the edges with two end nodes from each pair of adjacent class. A sequence of these node pairs along a boundary produces a visualization of the boundary. 

\begin{figure}[thb]
\centering
\subfloat[Decision boundary - Class 'bird' to Class 'airplane' (CIFAR-10).]{{\includegraphics[width=0.5\linewidth]{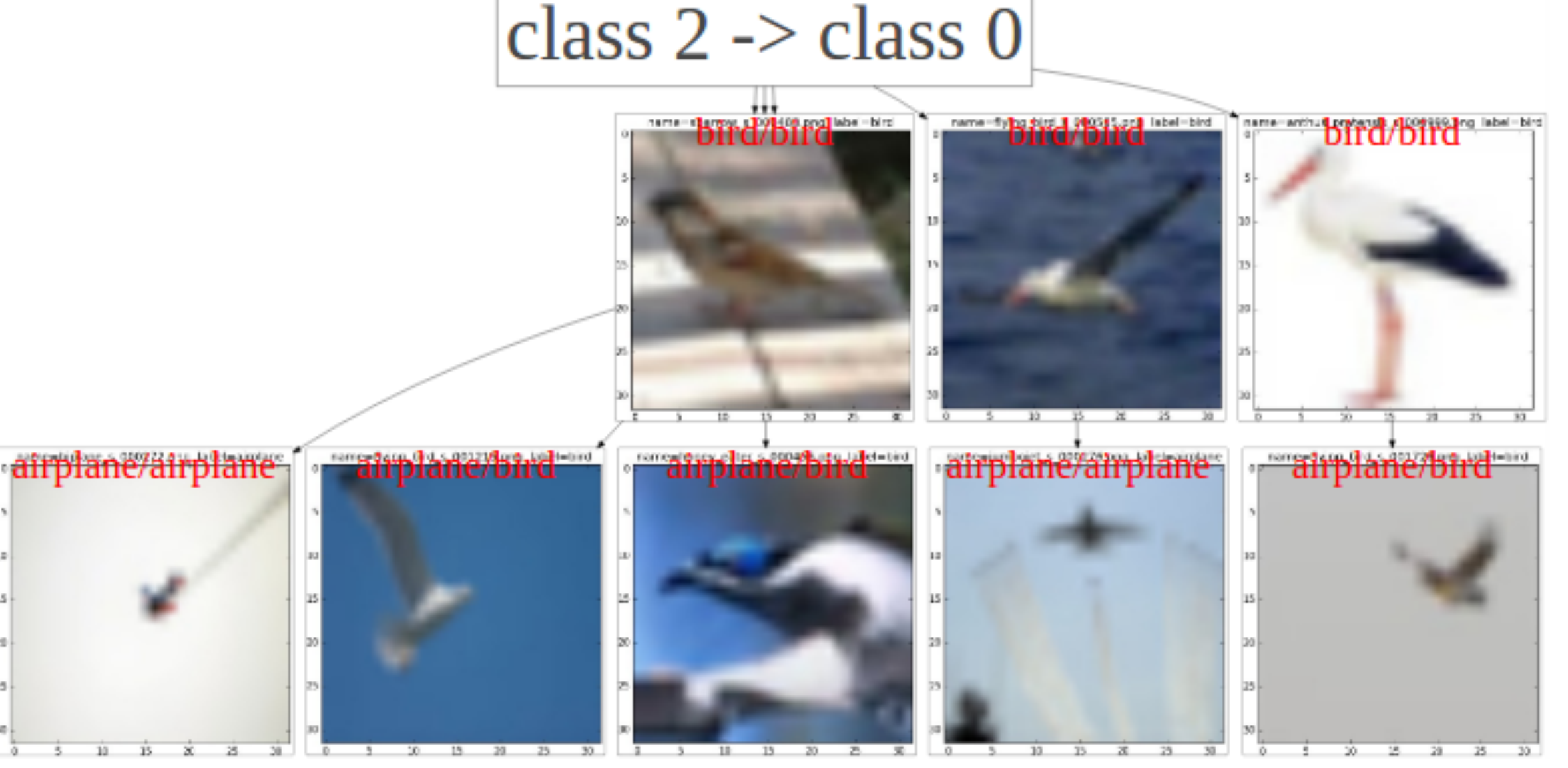}}
\label{fig:cifar_boundary_02}}
\hfill
\subfloat[Decision boundary - Class '2' to Class '7' (MNIST).]{{\includegraphics[width=0.4\linewidth]{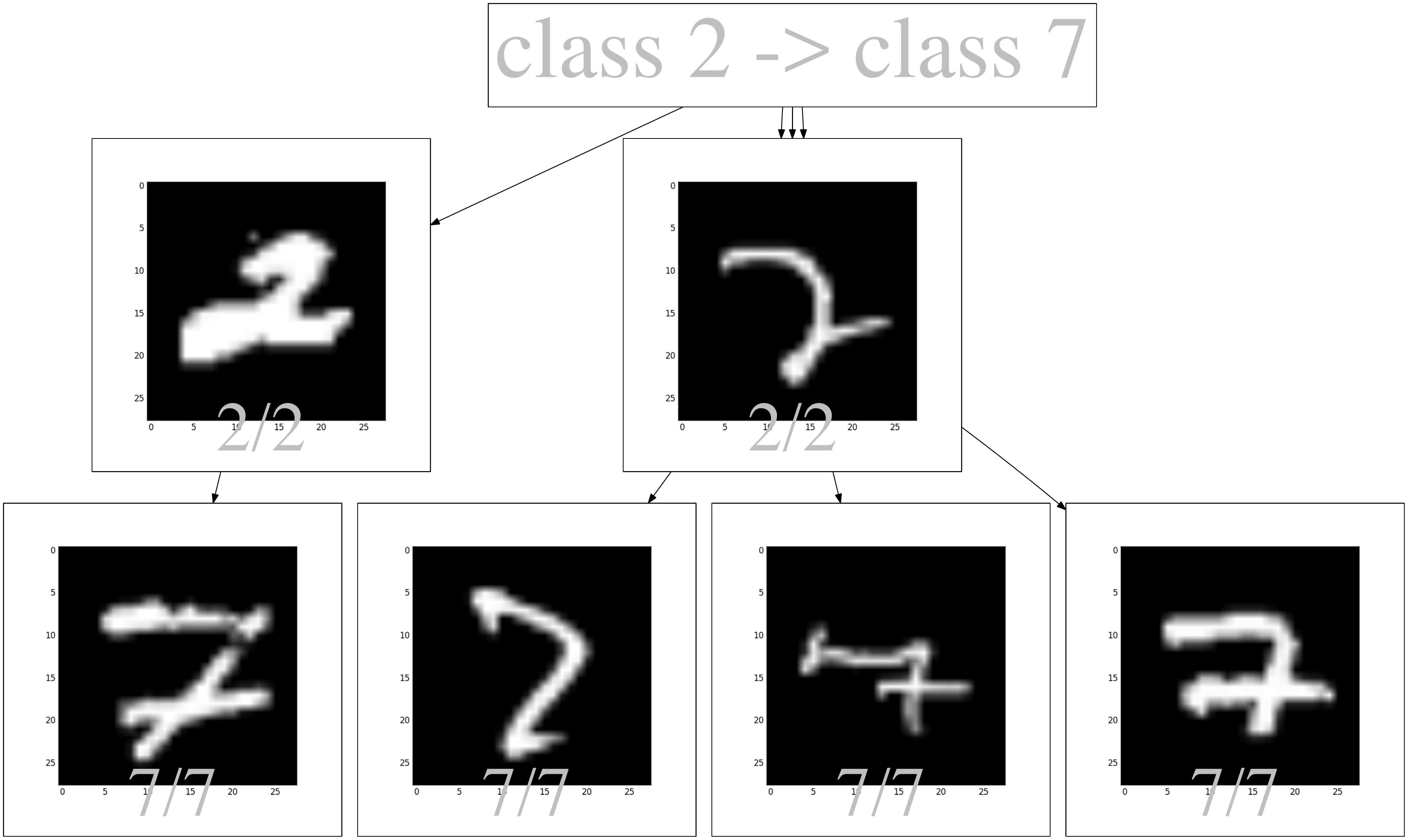}}
\label{fig:mnist_boundary_27}}
\caption{Classification boundary examples on CIFAR-10 and MNIST dataset.}
\label{fig:boundary_description}
\end{figure}

\begin{figure}[thb]
\centering
\subfloat[]{{\includegraphics[width=0.50in]{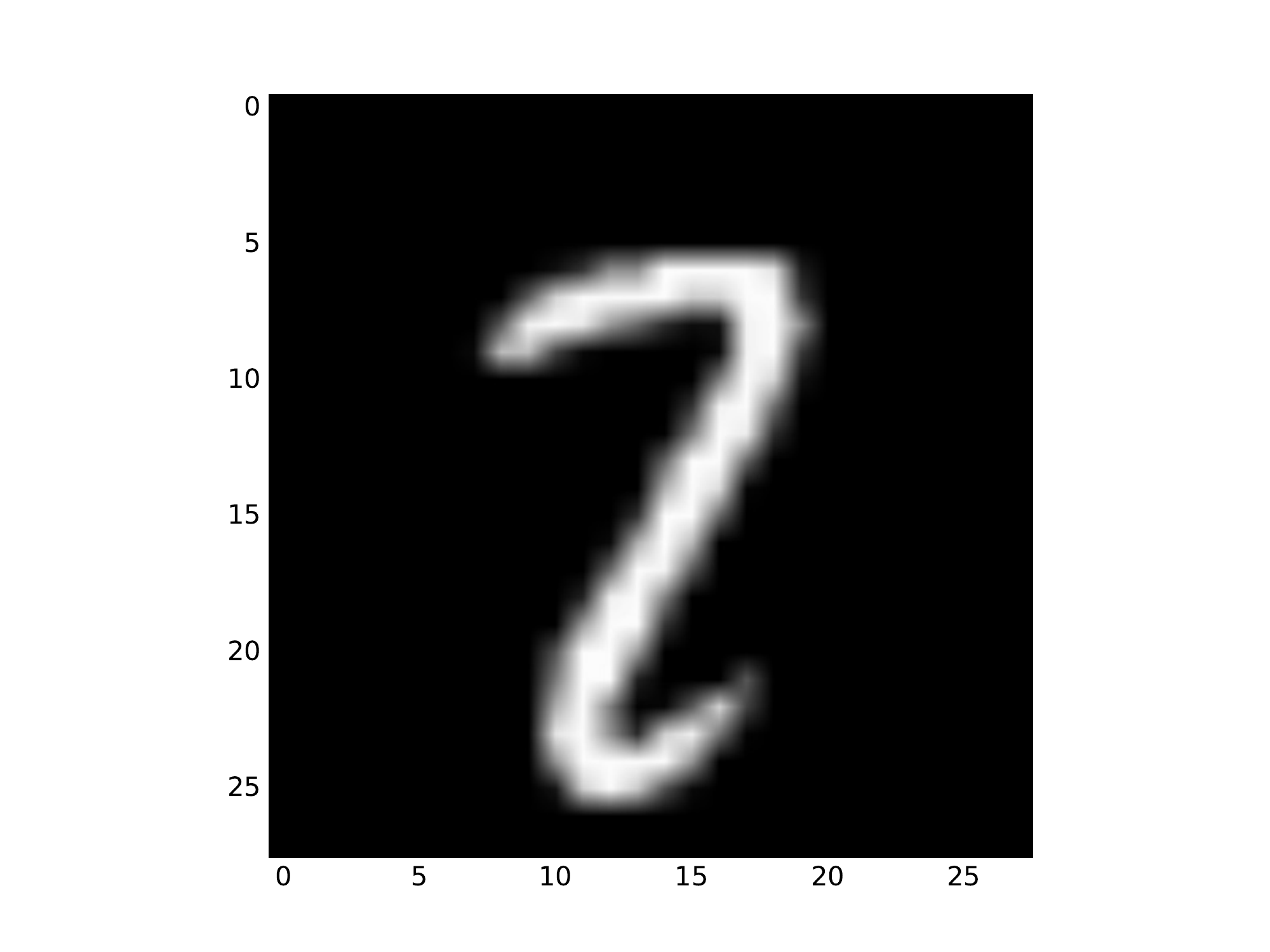}}
\label{fig:mnist_321}}
\hfill
\subfloat[]{{\includegraphics[width=0.50in]{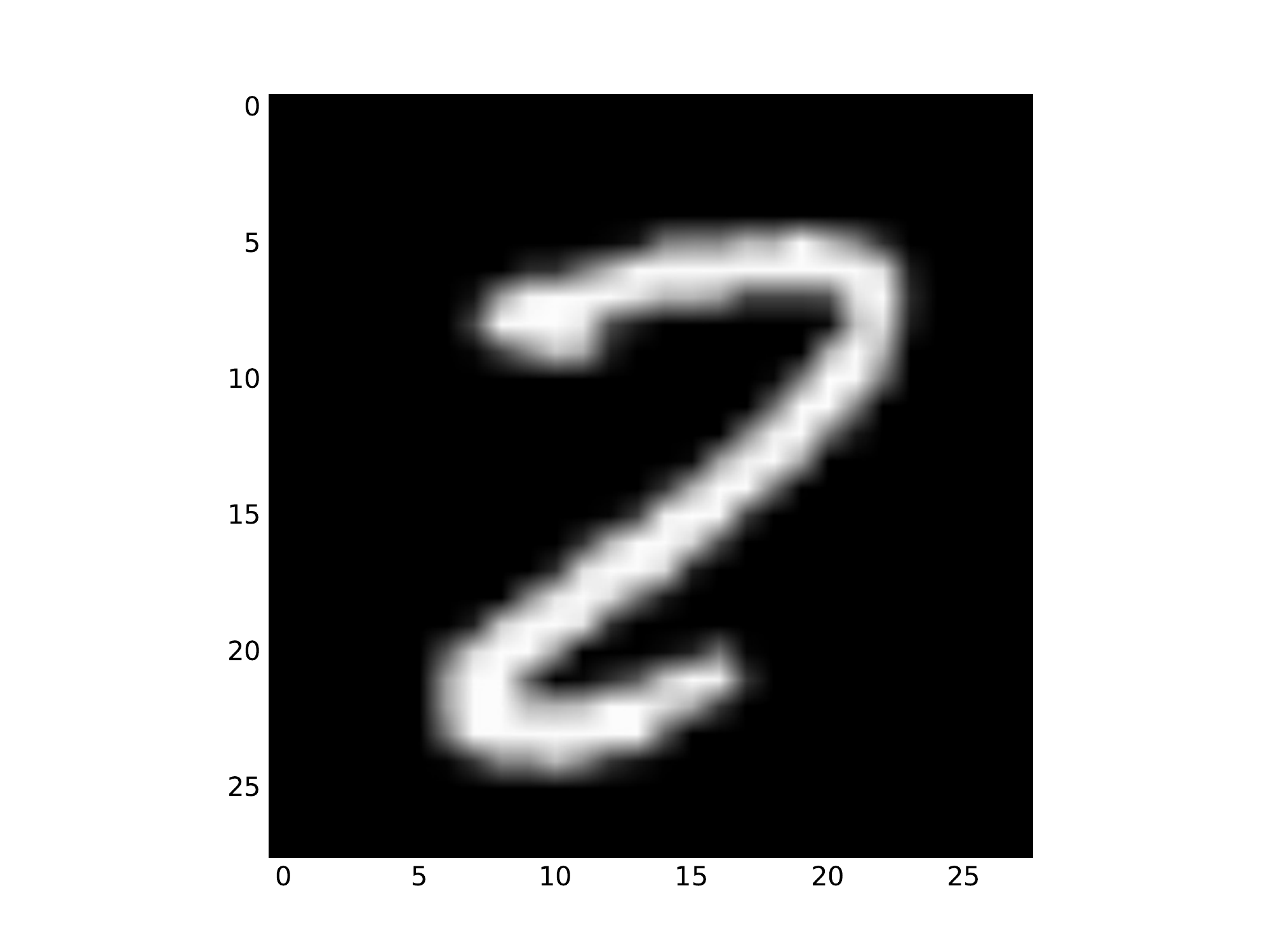}}
\label{fig:mnist_444}}
\hfill
\subfloat[]{{\includegraphics[width=0.50in]{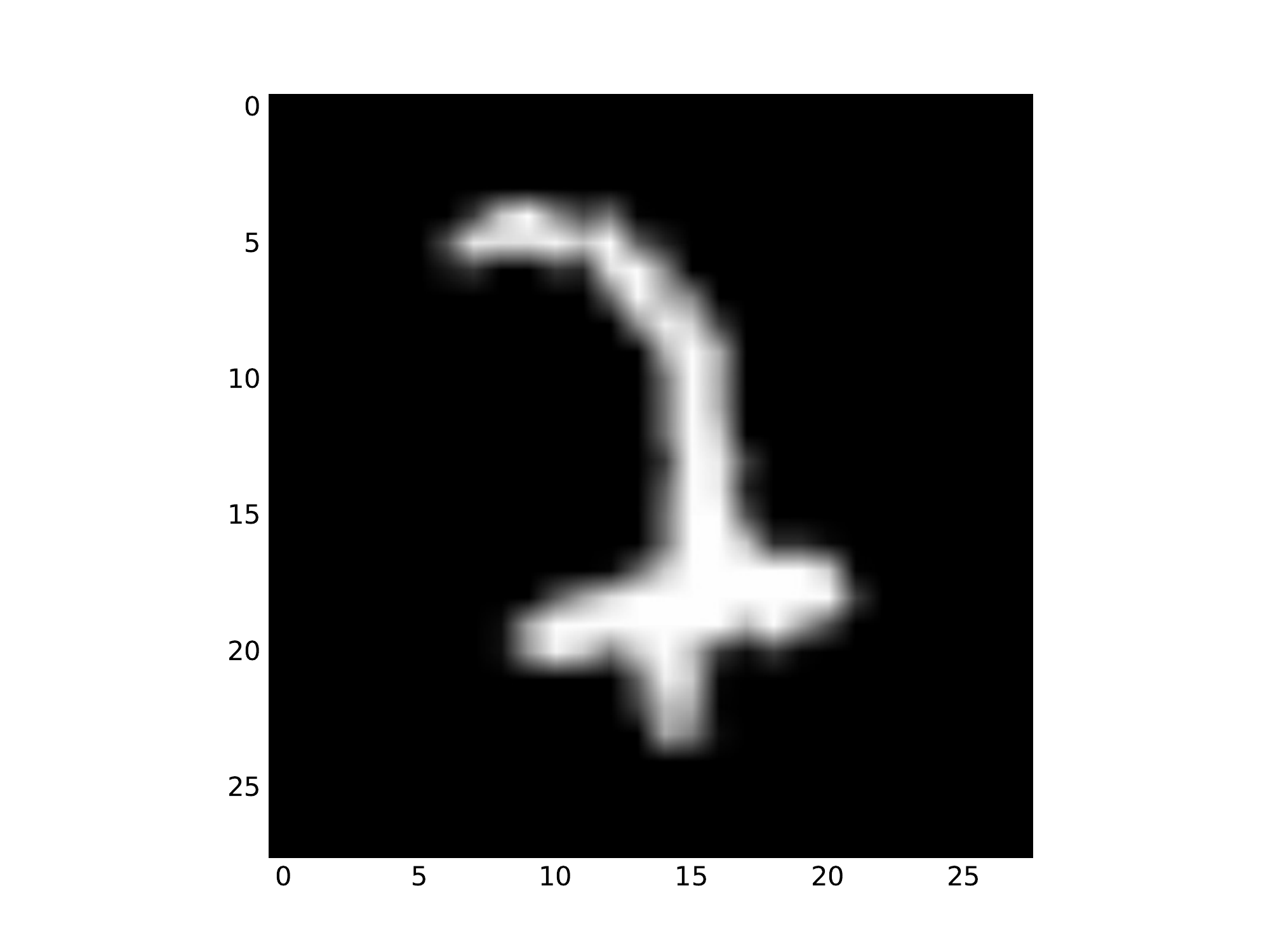}}
\label{fig:mnist_1226}}
\hfill
\subfloat[]{{\includegraphics[width=0.50in]{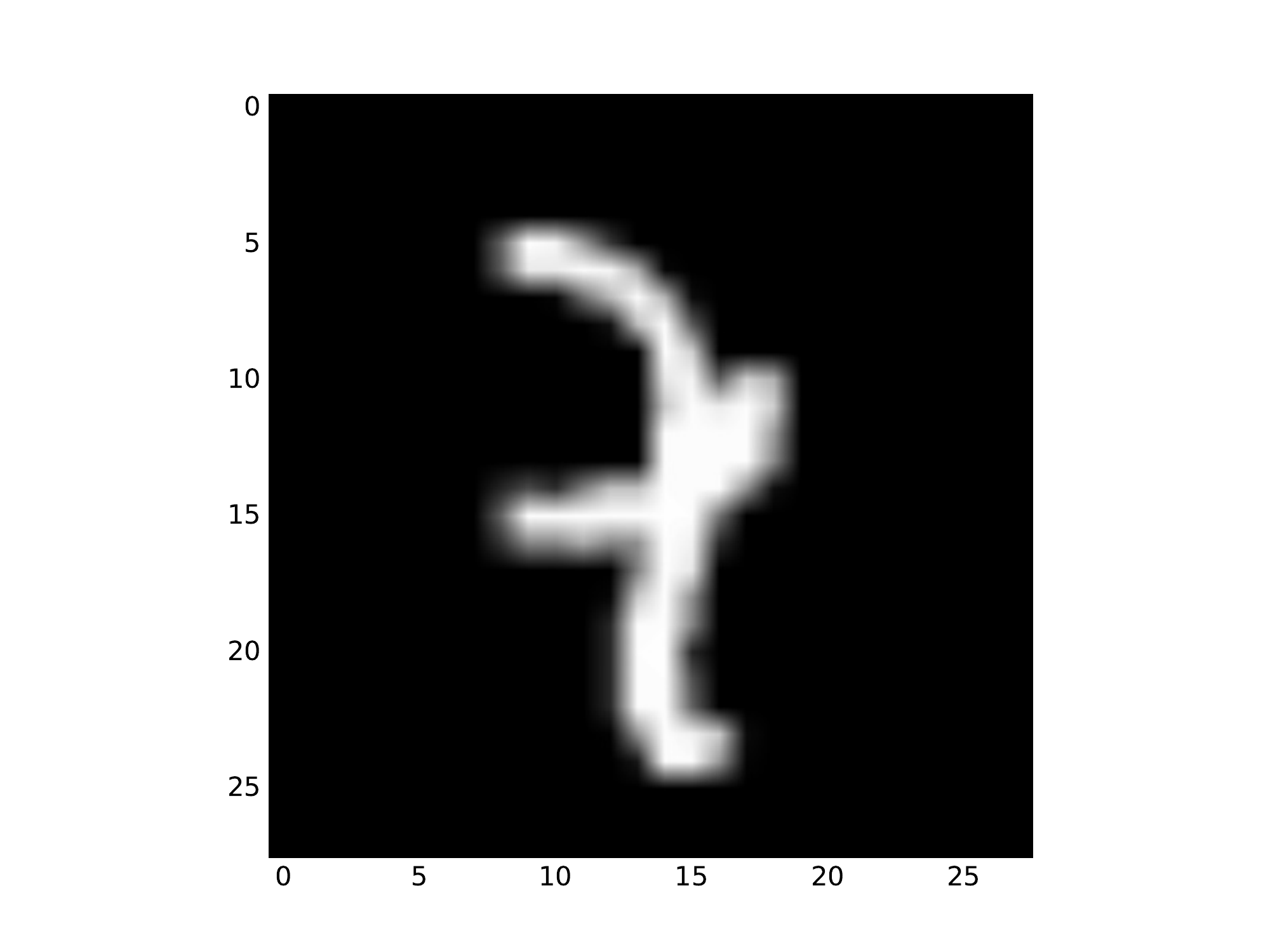}}
\label{fig:mnist_551}}
\hfill
\subfloat[]{{\includegraphics[width=0.33in]{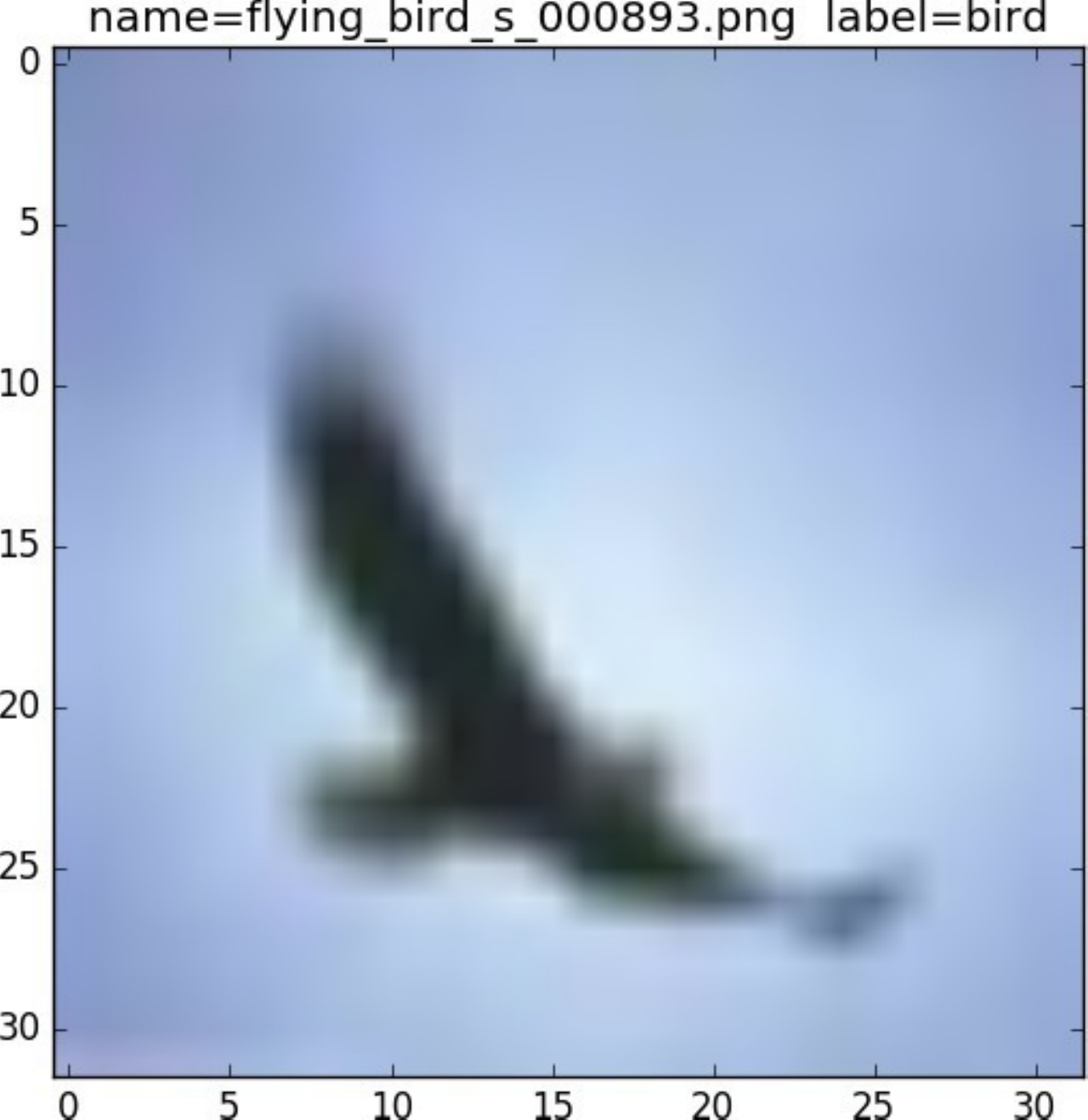}}
\label{fig:cifar_1495}}
\hfill
\subfloat[]{{\includegraphics[width=0.33in]{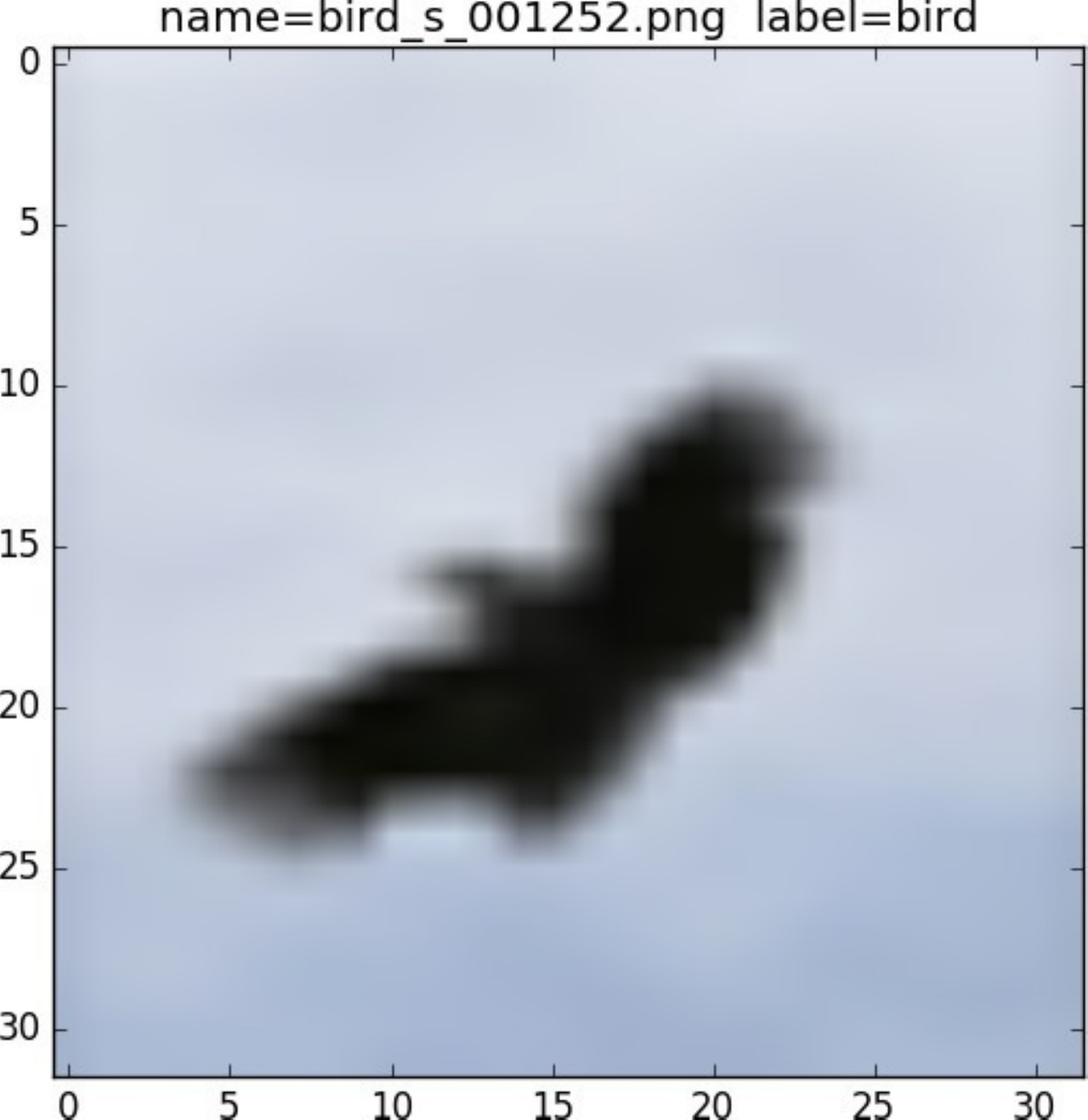}}
\label{fig:cifar_1547}}
\caption{Selected test samples in MNIST and CIFAR-10 dataset.}
\label{fig:selected_test_samples_boundary}
\end{figure}

By utilizing the boundary projections, model users can find some hints about subtle differences between two classes. To quantitatively analyze the effect, we present boundaries visualized this way to the 10 human subjects and ask them whether they understand decision making of the DNN model better with these boundaries. Participants may describe their understanding about how a model differentiates two classes in plain text. By looking at the decision boundary between class 2 to class 7 in MNIST dataset, as shown in Figure \ref{fig:mnist_boundary_27}, 90\% of the participants believe that they understand the decision making logic of the model. 

We further evaluate the observation made by human subjects by examining the classification difference with examples in training data. This is done as follows: It is noticed that the horizontal line at the bottom of a digit should be long enough for the digit to be classified into 2. If the horizontal line of a digit is short at the bottom or is in the middle, the digit is often classified as 7. To verify this observation, we selected several test samples in MNIST dataset to check their predictions. We choose four samples in MNIST dataset (see Figure \ref{fig:mnist_321} to \ref{fig:mnist_551}). As expected, the first and fourth are classified as 7 while the second and the third are classified as 2. 
Similarly, Figure \ref{fig:cifar_boundary_02} shows the boundary between 'bird' and 'airplane' for the CIFAR-10 dataset. One may note that some birds in training data are at the airplane side of the boundary (e.g., the bird at rightmost). This is caused by training errors. Our participants reach a conclusion that the birds with upwards wings are likely to be misclassified as airplanes. To verify, we choose test data with upwards wings in CIFAR-10 dataset (see Figure \ref{fig:cifar_1495} and \ref{fig:cifar_1547}) and check their predictions made by the DNN model. Not surprisingly, the DNN classifier classifies them into class 'airplane' incorrectly.

\begin{figure}[thb]
\centering
\subfloat[Decision boundary - Class 3 to Class 5.]{{\includegraphics[width=3.4in]{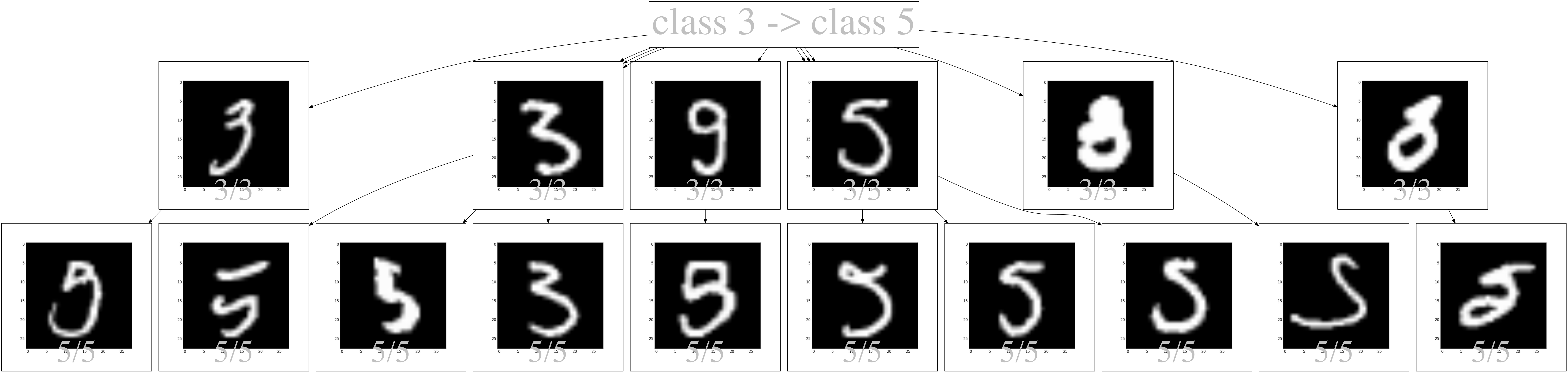}}
\label{fig:mnist_boundary_35}}
\vfill
\subfloat[Decision boundary - Class 5 to Class 3.]{{\includegraphics[width=3.4in]{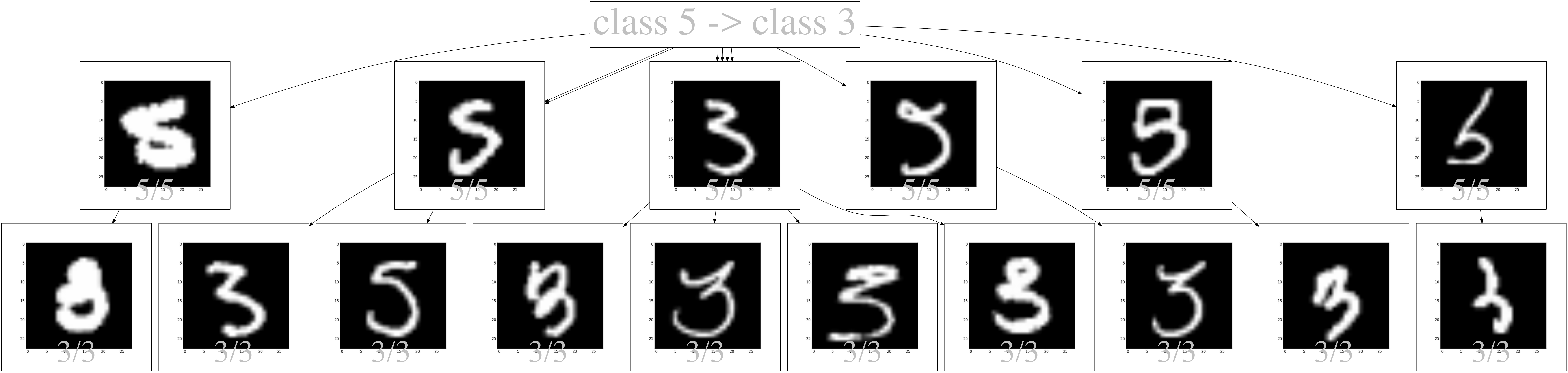}}
\label{fig:mnist_boundary_53}}
\caption{Classification boundary examples between class 3 and class 5 for LeNet-5 on MNIST dataset.}
\label{fig:mnist_boundary_description}
\end{figure}

The boundary projection also enables identifying incorrectly labeled training data through unexplainable observations. If a training sample of class B is mislabeled as class A, it still shares significant similarity with many samples in class B. The mislabeled sample can be identified near the boundary. As shown in Figure \ref{fig:mnist_boundary_35}, the label for the 4th sample is '3'. However, it is hard for one to guess why it is labeled as '5' rather than '3'. Similarly, the 3rd sample at the class 3 side and the 5th sample at the class 5 side both look like a 9 while they are labeled by '3' and '5', respectively. Our investigation afterwards confirms these samples are indeed mislabeled \footnote{http://deepmachinelearning.blogspot.com.au/2016/07/mnist-naughties.html}. These two samples also appear in class 5 to class 3 boundary (see Figure \ref{fig:mnist_boundary_53}).

\subsubsection{Case 3: Detecting Emerging New Classes.}
Detecting emerging new classes is a typical problem in streaming classification as predicting a sample from a new class not seen in training may lead to unexpected results. The challenge is that it requires massive distance computation~\cite{mu2017streaming,jordaney2017trasend} to compare the incoming samples with the existing training samples. EB-tree is able to reduce the computing complexity by only computing distances between a test sample and those training data points that have the same traversal path (and final node) with the test sample on the EB-tree. Formally, for an incoming sample $z$ that reaches node $N$ in EB-tree traversal and its predicted class $D$, we only need to compare a subset $D_{N}$ that reach node $N$ on the EB-tree through the same traversal path to detect new classes. As in~\cite{jordaney2017trasend}, we use conformal evaluation to compute the similarity between $z$ and data points in $D_N$ through p-values. The p-value $p_z^{D_N}$ of an incoming test data point $z$ is calculated as below:

\begin{equation} \label{concept_evo_eq1}
\alpha_{z} = A(D_{N},z)
\end{equation}
\begin{equation} \label{concept_evo_eq2}
\forall i \in D_{N}.\alpha_{i} = A(D_{N} \backslash z_{i},z_{i})
\end{equation}
\begin{equation} \label{concept_evo_eq3}
p_{z}^{D_{N}} = \frac{|\{j:\alpha_{j} \geq \alpha_{z}\}|}{|D_{N}|}
\end{equation}

In which, $A$ is a distance function. The distance between two data points that end their EB-tree traversal at node $N$ is defined as their probability distribution difference among node $N$,  the parent and children of node $N$. The probability distribution is calculated in a similar way as in~\cite{zoran2017learning}. The p-value $p_{z}^{D_{N}}$ for $z$ indicates how different the new data point is from existing data points that share the same traversal path. A low p-value indicates that the prediction on the test data lacks of statistical support for fitting the prediction model.

\begin{figure}[thb]
\centering
\subfloat[A sample 9.]{{\includegraphics[width=0.2\linewidth,valign=c]{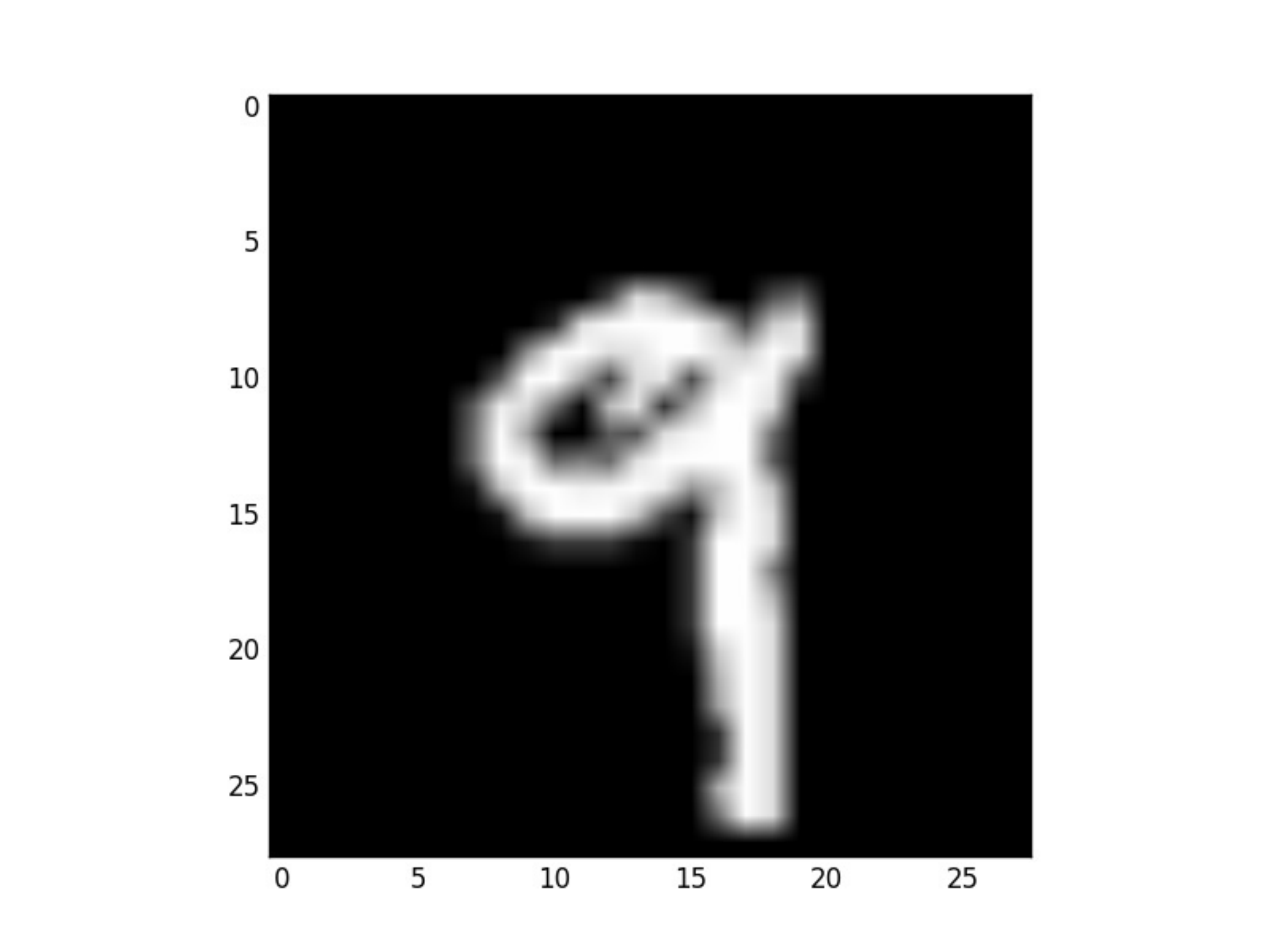}}
\label{fig:mnist_sample_9}}
\qquad
\subfloat[Traversal path for the sample.]{{\includegraphics[width=0.25\linewidth,valign=c]{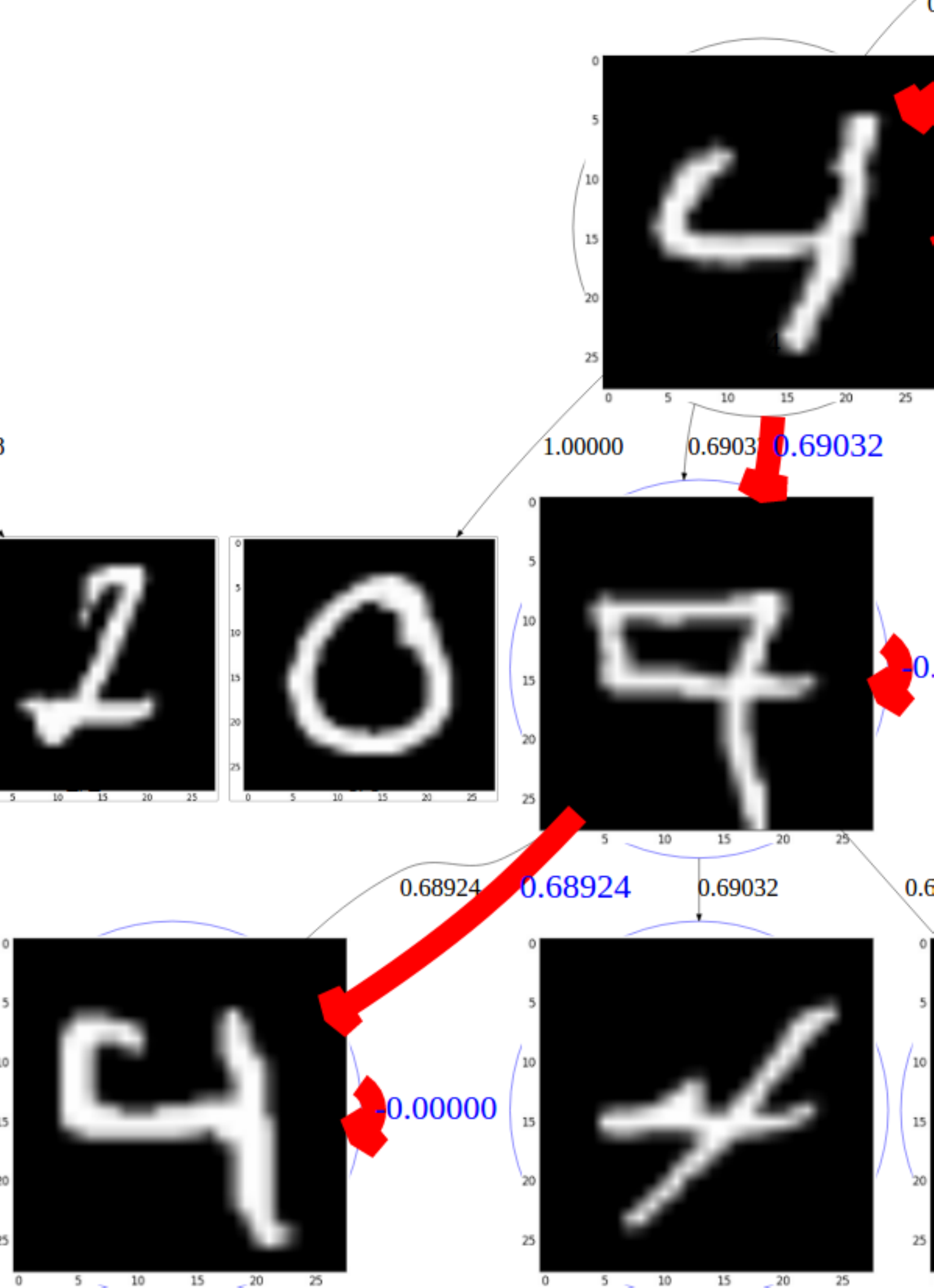}}
\label{fig:sample_traversal_path}}
\caption{A test image '9' is confidently classified as 4 on the LeNet which is only trained on data from 0 to 8. }
\label{fig:concept_evolution_detect}
\end{figure}

To demonstrate the effectiveness of EB-tree on emerging new class detection, we simulated a data stream with the MNIST dataset. LeNet-5 is used as the DNN classification model. Initially, the model is only trained by data from class 0 to 8. The EB-tree constructed for the DNN model contains 52 nodes. We then mix samples from class 9 in the stream and see whether they can be accurately identified. For a test sample S (see Figure \ref{fig:mnist_sample_9}) with true label 9, the sample is misclassified as a '4' because samples of class '9' have not been seen before. We compare all the nodes that reach the node that produces a prediction of '4' on the EB-tree and we get $p_{S} = 0.0069$. This value indicates the sample is statistically different to previous data points being classified as 4 through the same node and is likely to be from a new class. By setting a threshold of 0.01 for p-value, we correctly identified 991 out of 1000 samples (99.1\% accuracy) belonging to class '9' in the stream. Comparing to the existing method~\cite{jordaney2017trasend} that achieves an accuracy at 99.2\%, EB-tree reduces the computation by around 97.1\% through only computing the distances between an incoming sample and existing data points ending their EB-tree traversals on the same node. EB-tree provides a unique advantage for detecting emerging new classes and significantly reduces the detection cost.

\section{Conclusion}
We presented an Explicable Boundary Tree (EB-tree) to improve the trustworthiness of a shared complicated DNN model. EB-tree used a boundary stitching algorithm to equip the basic boundary tree with interpretability by disclosing a small set of representative training data. A model user gained insight of the decision making of a DNN model via tree traversal. We showed that an EB-tree approximated the corresponding DNN model with high fidelity and improved a model user's understanding of a complicated model. We also showed an EB-tree was effective on detecting mislabeled training data and emerging new classes in data.

\bibliographystyle{aaai}
\bibliography{moat}
\end{document}